\documentclass[letterpaper]{article} 
\usepackage{aaai24}  
\usepackage{times}  
\usepackage{helvet}  
\usepackage{courier}  
\usepackage[hyphens]{url}  
\usepackage{graphicx} 
\usepackage{natbib}  
\usepackage{caption} 
\usepackage{epsfig}
\usepackage{amsmath}
\usepackage{amssymb}

\usepackage[dvipsnames]{xcolor}
\usepackage{eso-pic}
\usepackage{xspace}
\makeatletter
\DeclareRobustCommand\onedot{\futurelet\@let@token\@onedot}
\def\@onedot{\ifx\@let@token.\else.\null\fi\xspace}

\def\eg{\emph{e.g}\onedot} 
\def\ie{\emph{i.e}\onedot} 
 
\def\etc{\emph{etc}\onedot}

\makeatother

\def\ul{\underline}
\usepackage{booktabs}

\usepackage{subcaption}  
\usepackage{multirow}
\newcommand\ourDS[1][]{LaSCo\xspace}
\newcommand\ourDSfull[1][]{Large Scale Composed Image Retrieval\xspace}
\newcommand\our[1][]{CASE\xspace}
\newcommand\ourfull[1][]{Cross-Attention driven Shift Encoder\xspace}

\newcommand\fashioniq[1][]{{{\it FashionIQ}}\xspace}
\newcommand\cirr[1][]{{{\it CIRR}}\xspace}
\newcommand\coir[1][]{{CoIR}\xspace}

\RequirePackage{cite}     

\usepackage[capitalize]{cleveref}
\crefname{section}{Sec.}{Secs.}
\Crefname{section}{Section}{Sections}
\crefname{figure}{Fig.}{Figs.}
\Crefname{figure}{Figure}{Figures}
\crefname{table}{Tab.}{Tabs.}
\Crefname{table}{Table}{Tables}
\usepackage{algorithm}
\usepackage{algorithmic}

\usepackage{newfloat}
\usepackage{listings}
\DeclareCaptionStyle{ruled}{labelfont=normalfont,labelsep=colon,strut=off} 
\floatstyle{ruled}
\newfloat{listing}{tb}{lst}{}
\floatname{listing}{Listing}
\title{Data Roaming and Quality Assessment for Composed Image Retrieval}
\author{
    Matan Levy\textsuperscript{\rm 1}
\qquad
Rami Ben-Ari\textsuperscript{\rm 2}
\qquad
Nir Darshan\textsuperscript{\rm 2}
\qquad
Dani Lischinski\textsuperscript{\rm 1}
}
\affiliations{
    \textsuperscript{\rm 1}The Hebrew University of Jerusalem, Israel\\
    \textsuperscript{\rm 2}OriginAI, Israel
}

\usepackage{bibentry}

\begin{document}

\maketitle
\begin{abstract}
The task of Composed Image Retrieval (CoIR) involves queries that combine image and text modalities, allowing users to express their intent more effectively. However, current CoIR datasets are orders of magnitude smaller compared to other vision and language (V\&L) datasets. Additionally, some of these datasets have noticeable issues, such as queries containing redundant modalities. To address these shortcomings, we introduce the Large Scale Composed Image Retrieval (LaSCo) dataset, a new CoIR dataset which is ten times larger than existing ones. Pre-training on our LaSCo, shows a noteworthy improvement in performance, even in zero-shot. Furthermore, we propose a new approach for analyzing CoIR datasets and methods, which detects modality redundancy or necessity, in queries.
We also introduce a new CoIR baseline, the Cross-Attention driven Shift Encoder (CASE). This baseline allows for early fusion of modalities using a cross-attention module and employs an additional auxiliary task during training. Our experiments demonstrate that this new baseline outperforms the current state-of-the-art methods on established benchmarks like FashionIQ and CIRR.

  \end{abstract}
\section{Introduction}
\label{sec:intro}
\begin{figure*}[t]
	\centering
	\includegraphics[width=0.9\linewidth]{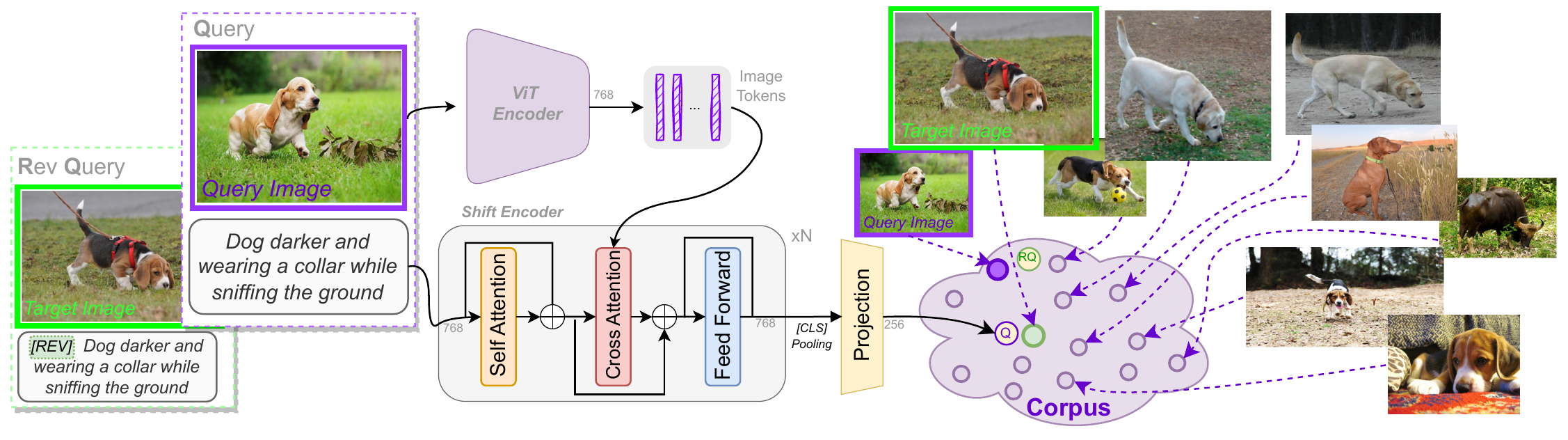}
	\caption{An overview of our \our baseline. The query-image is fed to a ViT encoder. The query-text is handled by our shift encoder \ie a BERT encoder with intermediate cross-attention layers, that receives the ViT output and fuses textual and visual information. The resulting $[\mathit{CLS}]$ token is then pooled and projected into a shared 256D space (circled $Q$). Finally, the $K$ nearest cosine-similarity neighbors of $Q$ are retrieved. For each training query, we also create a Reverse-Query by switching the roles of query and target images. A learnable special token $[\mathit{REV}]$ in our transformer, handles the {\it prediction of the query-image} (circled $RQ$) as the Reverse-Query task.}
	\label{fig:arc}
\end{figure*}

Recent progress in the field of multi-modal learning \cite{clip, vilbert} has been reflected in various downstream tasks, \eg, VQA \cite{VQA, levy2022classification}, Visual Dialog \cite{visdial}, Image captioning \cite{li2020oscar}, Image Retrieval (in its variations) \cite{cclip, Align_before_use} and Composed Image Retrieval (\coir) \cite{tirg, VAL_IR, cclip}.
Image Retrieval (IR) is a longstanding task that aims to find a desired image in a large corpus, given a user query. While content-based image retrieval uses a single visual modality to convey the user intent ~\cite{barz2021content, dubey2021decade, fewshotIR_ICIP2020}, providing a bi-modal query can mitigate miss-interpretations. In \coir
the gist and attributes are more succinctly described visually, and further intent is specified via a lingual modality \cite{fashion200k, MITstates, tirg, DialogBasedIIR, fashioniq, cirr, Arithmetic_multimodal_IR}. Some examples of \coir queries and their results are shown in \Cref{fig:arc,fig:fiq_cirr_ex}.

Despite the progress in foundation models and new \coir architectures, curating a dataset for CoIR remains a challenging chore, where the samples are triplets of query-image, accompanying transition-text, and the target image, serving as the ground-truth answer.
There are several existing datasets for \coir that differ significantly from each other. \citet{tirg} propose a dataset of rendered images of simple 3D scenes. Other existing datasets suffer from a small amount of data, and some are domain-specific (\eg, shoes \cite{DialogBasedIIR}), while in others, the lingual modality is limited by transition-text used as a class label \cite{MITstates}, or generated based on pairs of image captions that differ by a single word~\cite{fashion200k}. Another dataset was labeled based on previous vision and language (V\&L) models~\cite{Arithmetic_multimodal_IR}. Recently, \citet{fashioniq} introduced \emph{FashionIQ}, another domain-specific dataset for \coir, which gained popularity (\eg \cite{FashionVLP, Lee_2021_CVPR_cosmo, cirr}
) and contains human-annotated labels.

In addition to their small scale, shortcomings of these datasets include: 1) Not all acceptable target images for a given query are labeled as such, leading to incorrect count of false-negatives (\eg, \cref{fig:fiq_cirr_ex}); 2) Lack of visual complexity (due to restriction to a specific domain); and 3) Modality redundancy, \ie target images may often be retrieved using solely the query text, when descriptive enough to ignore the query image. We further refer to this issue as (lack of) {\it compositonality}: where the target should be determined by its query constituents combining the lingual and visual cues.

To break out from the previous domain-specific datasets to general and natural image domain, \citet{cirr} introduced the new \cirr (Composed Image Retrieval on Real-life images) dataset that contains open domain {\it natural images}, taken from NLVR2~\cite{nlvr2}. To the best of our knowledge, \cirr is the only existing dataset for \coir based on natural images with human annotated open-language texts. 
Despite attempts to reduce the false-negatives and relying on direct human labeling, \cirr still has two major shortcomings: 1) The image corpus is small, alleviating retrieval; 2) Modality redundancy still exists (see \cref{sec:modality_redundancy}), as well as false-negatives (according to our observations), reflecting the challenge in creating a ``clean" dataset.


In this work, we introduce a new large scale dataset for \coir, dubbed \ourDS (\ourDSfull dataset).
To construct it with minimal human effort, we employ a simple and effective methodology to rephrase labels from an existing large scale VQA dataset \cite{balanced_vqa_v2} into a form suited for \coir.
\ourDS contains an open and broad domain of natural images and rich text. Compared to \cirr, it has $\times\!10$ more queries, $\times\!2$ more unique tokens and $\times\!17$ more corpus images. We then propose a new approach for analyzing CoIR datasets and methods, which detects modality redundancy or necessity, in queries. Our analysis demonstrates that \ourDS shows a significantly smaller bias towards a single modality for retrieval. 

SoTA approach \cite{cclip} employ CLIP \cite{clip} to separately encode the textual and the visual query, followed by feature vector concatenation and a learnable projection head. We experiment with an additional end-to-end learnable baseline, which leverages the layers of BLIP's \cite{BLIP} image-grounded text encoder and enables early interactions between textual tokens and individual areas (patches) in the image. This baseline, dubbed \our ({\it \ourfull}), builds upon bi-modal {\it Cross-Attention} to create an {\it Encoder} that {\it Shifts} the query-image towards the target in the embedding space (see \cref{fig:arc}).
\our is trained using a novel bi-directional objective, which we refer to as {\it Reverse-Query} (RQ), where the query-image is predicted from the target-image and the query-text.
Being based on BLIP, \our uses a lower dimension latent vector of 256D, reducing retrieval complexity by a factor of $\times 2.5$ over previous SoTA. Furthermore, pre-training our baseline on \ourDS improves performance on \cirr dataset and even surpasses previous SoTA methods without training on \cirr (at zero-shot).

In summary, our key contributions in this paper are:
\begin{itemize}
\item {\it \ourDS:} A new large-scale, domain-free \coir dataset, a few orders of magnitude larger than existing datasets.
\item {\it Data Roaming:} A simple methodology for automatically generating \coir triplets from an existing VQA dataset. 
\item {{\it Modality Redundancy:} A method for analyzing redundancy between modalities in existing \coir datasets.}
\item {\it \our:} A new BLIP-based baseline, featuring early fusion and a novel bi-directional training objective, that achieves SoTA performance with a large gap on \fashioniq, \cirr and \ourDS benchmarks.
\end{itemize}
\section{Related Work}
\label{sec:related_work}

\paragraph{Data Roaming:}
A major challenge in many multi-modal tasks, such as text-video and text-audio retrieval, is the lack of large-scale training data. Due to the complexity involved in creating multi-modal datasets such as text to image \cite{flickr, coco} or video retrieval \cite{MSR_VTT, MSVD}, several studies suggest using raw narrated video footage \cite{howto100M} for video retrieval or altering the narration to create a dataset for Visual Question Answering (VQA) \cite{justAsk_VQA_2021}. Other works try to enhance existing datasets, \eg, COCO captioning \cite{nocaps2019} to more diversity and object categories. In this line of work, \citet{nagrani2022learning} propose a new video mining pipeline which involves automatically transferring captions from image captioning datasets to video clips, to create a new large-scale, weakly labelled audio-video captioning dataset. Nevertheless, for \coir models to ever function in the wild, a much larger variety of visual concepts must be learned, ideally from less annotated datasets. In this paper (\Cref{sec:lasco}) we propose a methodology for leveraging VQA2.0 \cite{balanced_vqa_v2}, a large existing and labeled dataset for the VQA task.

\coir datasets consist of triplets of query image, transition text and a target image. In order to differentiate these datasets from text-to-image and image-to-image retrieval, these triplets should ideally satisfy a condition where reaching the target image in the corpus will necessarily require both modalities. 
In this paper we further suggest an analysis tool for the ``quality'' of certain dataset, measured by ``modality redundancy''. We show that our newly generated large scale dataset exhibits higher quality, compared to CIRR on natural images, and is on-par with the domain-specific manually annotated FashionIQ dataset.

\paragraph{Composed Image Retrieval:} 
CoIR methods commonly learn a shared embedding space between the text and visual modalities. These methods often differ by encoding models, \eg, \cite{tirg} that uses ResNet and LSTM and learns a shift encoder. Other methods suggest different attention mechanisms, \eg, \citet{VAL_IR,HosseinzadehW20}. \citet{FashionVLP} focuses on specific domain characteristics such as the fashion domain (with FashionIQ dataset). Different fusion strategies between visual and textual modalities has gained high attention suggesting early \cite{HosseinzadehW20} and late \cite{ARTEMIS} fusion methods. 

Recent works leverage VLM foundation models, \eg \citet{cclip, Arithmetic_multimodal_IR} use CLIP features reaching top performance. 
Encouraged by \cite{cclip} we suggest a strong baseline built from pretrained BLIP components, finetuned on CoIR task.

Lastly, \citet{Kim_Yu_Kim_Kim_2021_dcnet} suggested enforcing cyclic consistency from query/target images back to the transition-text. To this end, they jointly optimized two separate networks, one devoted to the \coir task; another predicting the query text, given the query and target images. The latter is used for re-ranking the target candidates. In this work, we suggest a different auxiliary task, dubbed \emph{reverse objective}, inspired by the CoIR task.
Our reverse objective maps the target image, conditioned on the query text, back to the query image.

\section{\ourDS Dataset}
\label{sec:lasco}
In this section we introduce \ourDS ({\it \ourDSfull}), a new \coir dataset consisting of open-domain natural images, that elevates the scale of existing datasets.
To construct \ourDS, we leverage the carefully labeled datasets that exist for the well-studied VQA task \cite{justAsk_VQA_2021}. 
Specifically, we utilize the {\it VQA2.0} \cite{balanced_vqa_v2} dataset to create \ourDS with minimal human effort. 
{\it VQA2.0} introduces two important features: 1) A balanced answer set for VQA; 2) Inclusion of ``complementary'' samples, with counter examples.
A complementary image $I_c$ is one that is similar to an original image $I$ in VQA, but yields a different answer for the same question.

\begin{figure}[t]
  \centering
  \includegraphics[width=1\linewidth]{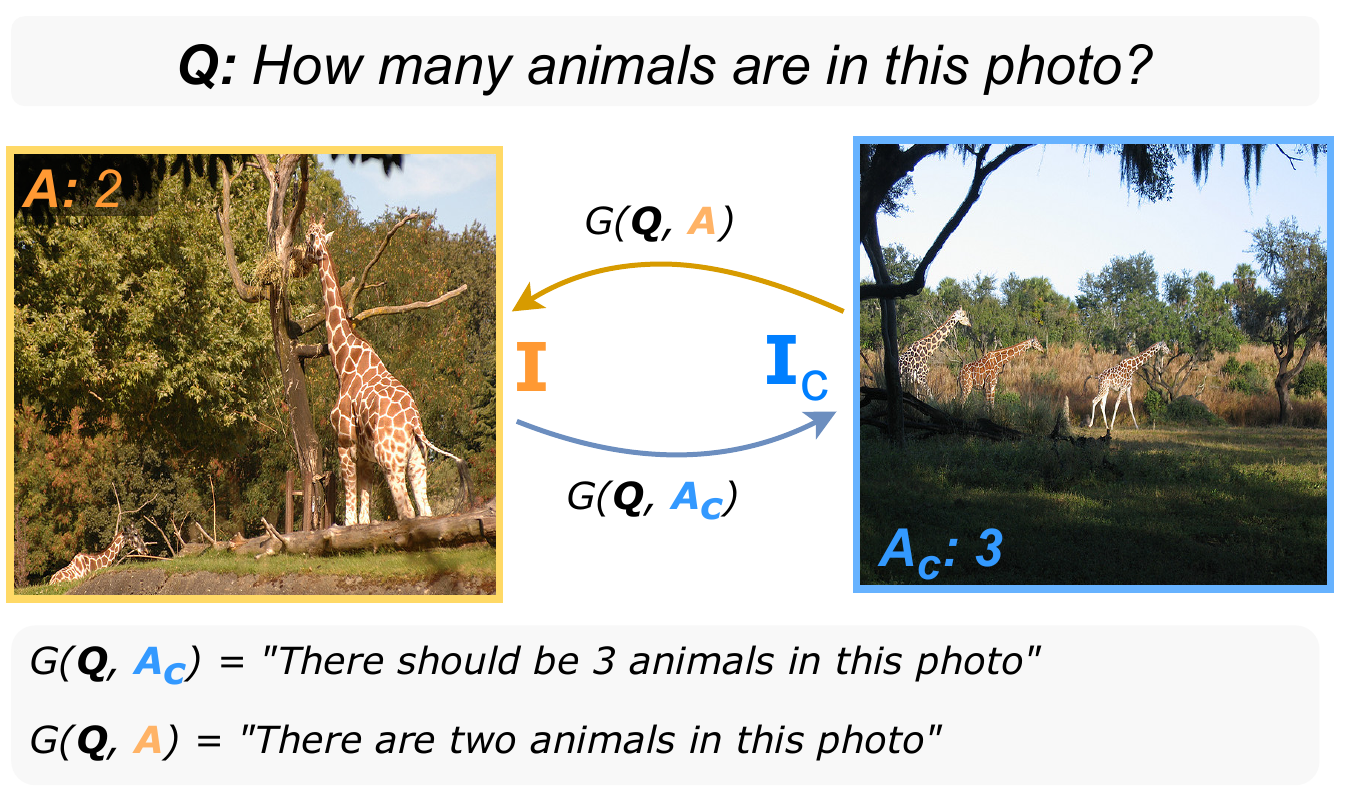}
   \caption{Generating transition texts from \textit{VQA2.0} samples. Given two paired triplets $(I, Q, A), (I_c, Q, A_c)$, where $I$ and $I_c$ are visually similar, but yield different answers $A$ and $A_c$ for the same question $Q$, a transition text from $I$ to $I_c$ is generated by $G:=${\it GPT-3} \cite{GPT_3}, based on $(Q, A_c)$.}
   \label{fig:our_vqa2}
\end{figure}

\subsection{Data Roaming}
\label{sec:Data Collection}
We generate \coir triplets from {\it VQA2.0} samples and their ``complementary'' counterparts, as demonstrated in \Cref{fig:our_vqa2}. 
For brevity, let us denote {\it VQA2.0} by set $\mathcal{D}$.
Consider two complementary triplets $(I, Q, A)\in \mathcal{D}$ and $(I_c, Q, A_c)\in \mathcal{D}$. By construction, each $I_c$ image was manually selected from the 24 (visually) nearest neighbors of $I$ such that: 1) any premise assumed in $Q$ must hold in $I_c$; 2) it makes sense to ask $Q$ about $I_c$; and 3) the answer $A_c$ to $Q$ about $I_c$ differs from $A$. The actual answer $A_c$ was provided by a different annotator, in a second round. 
These properties of $\mathcal{D}$ enable building our new dataset by using the existing $(Q,A_c)$ pair to construct the transition text $S_c$ from $I$ to $I_c$.

{\bf Transition text:} 
The text conversion task is defined by $G:(Q,A_c) \mapsto S_c$. We employ a strong language model, \eg, $G:=${\it GPT-3} due to its few-shot capabilities~\cite{GPT_3}, to perform the conversion. To create an example set for the task, we recruit 20 annotators to manually rephrase $\sim\!300$ randomly sampled $(Q,A_c)$ pairs to valid transition texts $S_c$. We then provide $G$ a short description of the task, with three annotated examples of $(Q,A_c,S_c)$, and ask the model to perform the task on a new pair $(Q,A_c)$. We further exploit the symmetry in the transition, $I\rightarrow I_c$ and $I\leftarrow I_c$ to generate more triplets. Finally, we organize the triplets as $(I,S_{c},I_c)$, $(I_c,S,I)$, with $S_{c}$ and $S$ indicating the corresponding transition texts. For an extensive list of examples we refer the reader to our suppl. material.

\subsection{Quality control}
We further conduct a data curation process and preliminary evaluation for the quality of our generated \coir triplets. We first remove triplets that share the same query/target image. For text, we apply automatic rule-based scripts to filter out potentially wrong conversions (\eg, too short, unexpected characters such as `\textbackslash{}n', etc.). In our manual examination of $1000$ random text conversions we judged $91.7\%$ of them to produce well-phrased and reasonable transition texts. Next, we conduct a short user study to compare the quality of our transition texts to fully human-annotated ones. We sample $\sim300$ random triplets $(Q_i, Q_t, T_i)$, of query-image, query-text, and target image (respectively). Triplets were shown to a total of 30 users, who were asked whether $Q_t$ ``adequately describes the transition/modification'', from $Q_i$ to $T_i$, or not. This experiment was conducted on three different CoIR datasets. The results show $82.02\%$ positive rate for \ourDS samples, compared to $81.15\%$ for \fashioniq, and $82.65\%$ for \cirr.

Finally, we performed a larger scale user study using the Amazon Mechanical Turk (AMT) platform. We randomly presented 1000 samples (triplets) from each dataset, and asked 3 different AMT workers to rate each sample using a 1–5 rating scale (worst–best, respectively). A mean opinion score (MOS) was computed for each sample as the average of the three ratings. Binarization of the ratings (considering 1,2 as ‘Bad’, otherwise as ‘Good’) yields a positive (Good) rate of 90.9\%, 93.8\% and 97.1\% for LaSCo, FashionIQ and CIRR, respectively. The overall (relative) gap between LaSCo and the other datasets is under 7\%, indicating that the generated texts are on-par with human annotations. For further information, please see our suppl.~material.

\begin{table}[ht]
\begin{center}
\resizebox{\columnwidth}{!}{%
\begin{tabular}{lcccccc}
\hline
Dataset & $\uparrow$\#Triplets & \begin{tabular}[c]{@{}c@{}}$\uparrow$\#Corpus\\ (Train.)\end{tabular} & \begin{tabular}[c]{@{}c@{}}$\uparrow$\#Corpus\\ (Val.)\end{tabular} & \begin{tabular}[c]{@{}c@{}}\#Unique\\ Tokens\end{tabular} & \begin{tabular}[c]{@{}c@{}}Avg. Txt.\\ Length\end{tabular} & \begin{tabular}[c]{@{}c@{}}Image\\ Domain\end{tabular} \\ \hline
CIRR & 36,554 & 16,742 & 2,297 & 6,880 & 59.51 & Natural \\
FashionIQ & 30,132 & 25,133 & 5,138 & 4,425 & 27.13 & Fashion \\
LaSCo (ours) & 389,305 & 81,653 & 39,826 & 13,488 & 30.70 & Natural \\ \hline
\end{tabular}
}
\end{center}
\caption{Comparison of \ourDS to existing Composed Image Retrieval (\coir) datasets, CIRR and FashionIQ.}
\label{tab:datasets}
\end{table}

\Cref{tab:datasets} compares statistics of \ourDS to previous \coir datasets. \ourDS contains over 389K queries, $\times$10 larger than previous datasets, with an image set containing 121.5K different images, compared to previous 19K--41K. The size of the test image corpus, determining the target search space, is almost 40K, compared to 2.3K in \cirr and 15.4K in \fashioniq. In terms of natural language, \ourDS is richer with 13.5K different language tokens, compared to 4.4K in \fashioniq and 6.8K in \cirr. Moreover, \ourDS and \emph{VQA2.0} are both derived from COCO's image set; thus, captions are available for each of \ourDS's images. Utilizing captions as an additional cue (see \Cref{sec:evaluation}) allows creating a rich dataset for training \coir methods to achieve high performance in both Text-to-Image and \coir tasks.

\section{\ourfull}
\label{sec:method}
Here we introduce a new strong baseline for \coir that leverages pre-trained BLIP components with early fusion, named
{\it \ourfull (\our)}.

\subsection{\our Architecture}
\label{sec:method_our}
The \our architecture, depicted in \Cref{fig:arc}, consists of two transformer components \cite{vaswani2017attention}. The first is a shift-encoder, based on an image-grounded text encoder, previously introduced in \cite{BLIP}. It is a BERT~\cite{devlin2019bert} encoder with additional intermediate cross-attention layers, to model vision-language interactions. The second component is a ViT \cite{DosovitskiyB0WZ21} encoder. ViT divides an input image into patches and encodes them as a sequence of \emph{image tokens}.
The image tokens are then fed into cross-attention layers, allowing interaction between the lingual and visual branches. The output, a bi-modality conditioned sequence (text on image and image on text), is then pooled to a single vector and projected to a 256D latent space. \our allows {\it early fusion} between modalities, in contrast to previous {\it late fusion} methods \cite{cclip, tirg, Kim_Yu_Kim_Kim_2021_dcnet, ARTEMIS} or methods that take a middle way \cite{cirr, FashionVLP, VAL_IR, HosseinzadehW20}, as discussed in \Cref{sec:related_work}.

{\bf Utilizing Vision-Language Pre-training:} We initialize our model's weights using BLIP  \cite{BLIP} pre-trained weights, as follows:
Our shift-encoder's Self-Attention, Cross-Attention and Feed-Forward layers
are initialized with the corresponding layers of BLIP's image-grounded encoder. Our final projection layer is initialized with the final projection of BLIP's text-encoder. Finally, we initialize our ViT component with BLIP's image encoder. Our model is end-to-end trainable (see \Cref{fig:arc}).
\subsection{Adding a Reverse Objective}
\label{sec:method_reverse_queries}
A common training approach for image retrieval tasks uses an objective of contrastive loss with the target image as positive \eg, \cite{FashionVLP, cclip, cirr, li2020oscar}.
Here, we propose an additional \emph{reverse} objective, where the goal is to retrieve the query image given the transition-text and the target image. One can view the reverse objective as flipping the shift vector of the original query (in the latent space) to point in the opposite direction, from the target image to the embedding of the query image.
Our reverse objective further suggests an additional task and can be viewed as a valid augmentation, effectively enlarging the dataset. We therefore train our model jointly with the reverse objective (see \Cref{fig:arc}). Namely, given a triplet $(Q_i, Q_t, T_i)$ of query-image, query-text and target-image (respectively), our model objective $M$ requires: $M(Q_i, Q_t)=T_i$ (standard \coir task), while simultaneously enforcing $M(T_i, [REV];Q_t)=Q_i$, where $[REV]$ is a special token provided to the model.
Although the reverse task is not one-to-one (multiple queries may be suitable), this objective has proven to be beneficial in practice.

\subsection{Retrieval Approach}
 We follow the most common approach for image retrieval: searching for matches in an embedding space shared by queries and targets (see \Cref{fig:arc}). First, corpus images (potential targets) are encoded by a ViT \cite{DosovitskiyB0WZ21} encoder, resulting in a single feature vector, representing each image. Then, a given query (composed of image and text) is projected by \our to  the shared embedding space. Finally, the target candidates are ranked by cosine-similarity distance w.r.t the shifted query embedding. By using a relatively small embedding space dimension of 256D, compared to 640D in the previous SoTA \cite{cclip}, the retrieval is sped up by $\times$2.5.
 
\section{Modality Redundancy}
\label{sec:modality_redundancy}

In this section, we first propose a simple analysis of existing \coir datasets to examine the degree to which their queries require \emph{both} modalities for successful retrieval. Next, a similar analysis is proposed for assessing the bias of \coir methods towards modality redundancies.

An ideal composed query should require both modalities for retrieving the desired target. For example, a transition-text such ``Change the color to be more cream colored'' in the top row of \Cref{fig:fiq_cirr_ex}, will only succeed in finding the proper target in conjuction with the query image, since the type of object cannot be inferred from the text alone.
However, in practice, one of the modalities can become redundant, with the degree of redundancy depending on the information conveyed by the other modality.
On one extreme, the query-image might be completely redundant, reducing the task to Text-to-Image retrieval; on the other extreme, the query-text might be redundant, with the task becoming Image-to-Image retrieval.
To quantitatively assess the degree of redundancy in \coir datasets, we measure the Recall@K performance of naive Text-to-Image and Image-to-Image retrieval using the embeddings produced by an independent off-the-shelf CLIP model \cite{clip}.
\begin{figure}[t]
  \centering
  \includegraphics[width=0.85\linewidth]{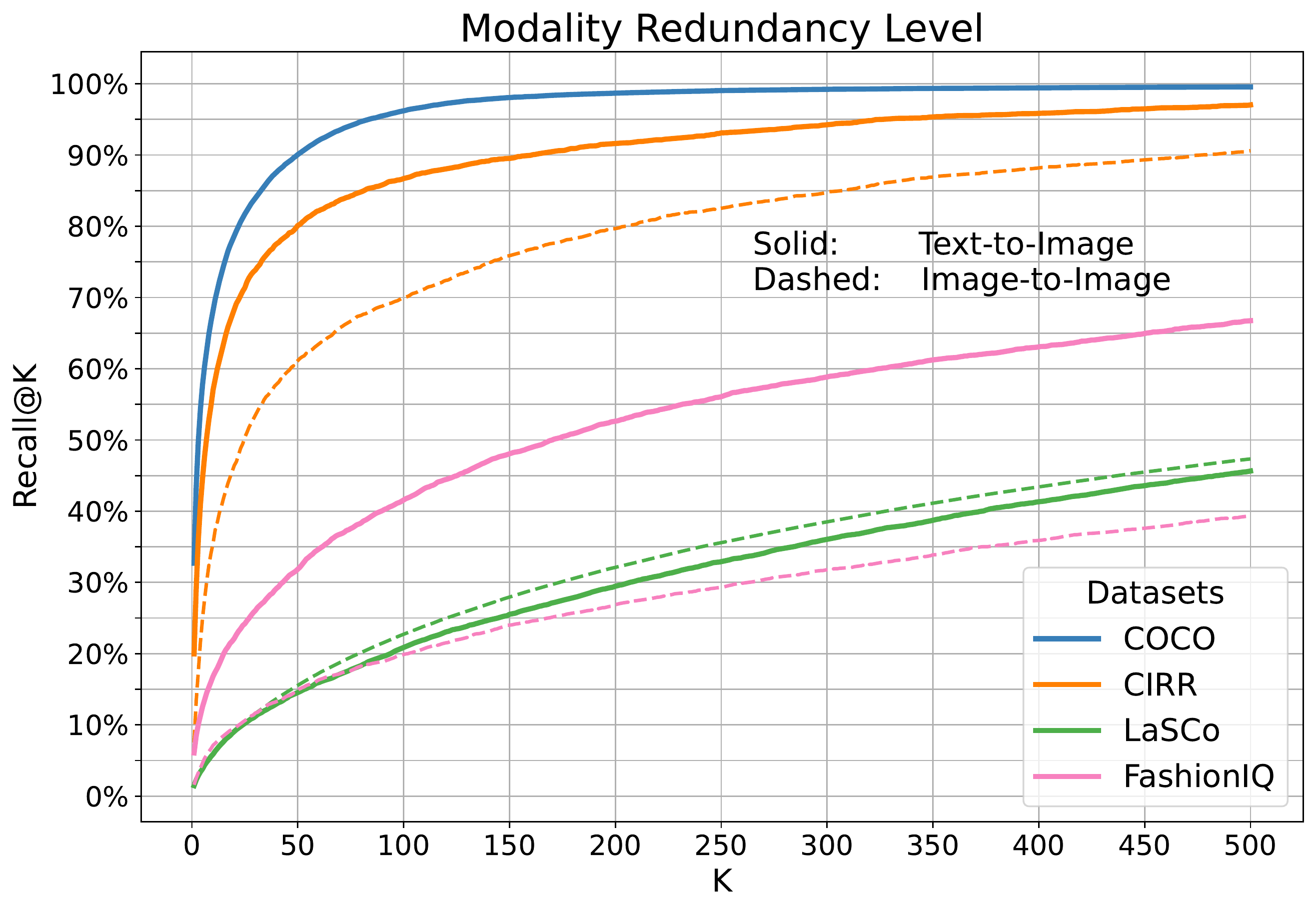}
   \caption{Modality redundancy level in several datasets (lower is better). Recall@K values for {\it Text-to-Image} and {\it Image-to-Image} retrieval using off-the-shelf CLIP \cite{clip}. Lower values indicate higher image and text compositionality. The results for  COCO  \cite{coco} are shown as a reference for a purely text-to-image (single modality) retrieval task, presenting an upper bound.
	}
   \label{fig:tti_clip}
\end{figure}
To create a continuous measure we compute the Recall@K for varying $K$ values. These measurements computed on several datasets are plotted in \Cref{fig:tti_clip}.
A lower curve indicates that the corresponding dataset is more challenging for a uni-modal query. Note that the \ourDS and \fashioniq curves are much lower than \cirr, implying that more of the queries in \cirr are modality-redundant.
For reference, we also plot the performance of the CLIP-based Text-to-Image retriever using COCO captions as query text (a commonly used benchmark for Text-to-Image retrieval \cite{BLIP, Align_before_use, li2020oscar, clip}). While COCO may be viewed as an ``upper bound'' for this task, note that the \cirr curve is quite close to it.

\begin{figure}[t]
  \centering
  \includegraphics[width=0.85\linewidth]{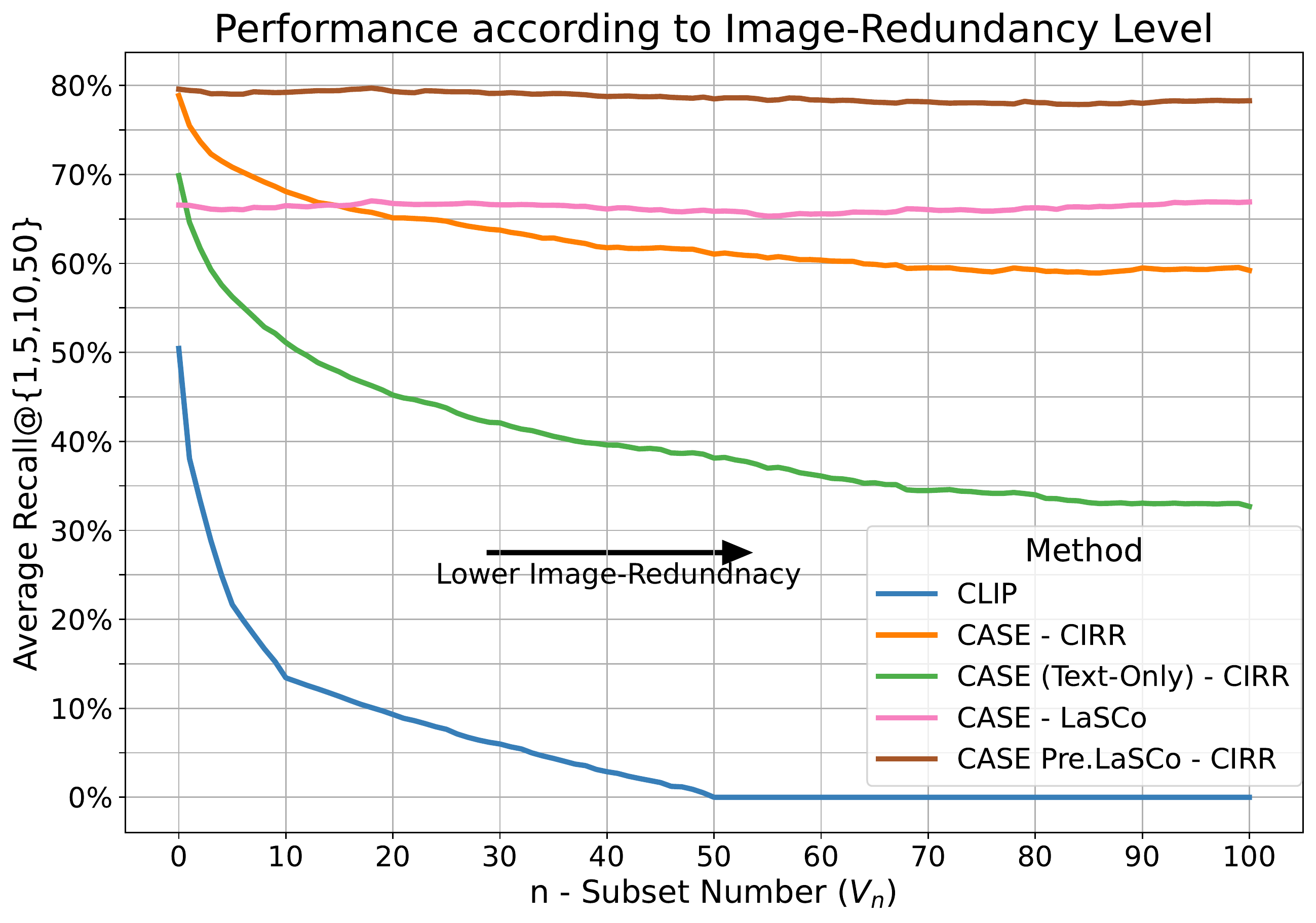}
       \caption{Average retrieval performance on subsets of \cirr determined by {\it image-redundancy} levels. Higher values of Average Recall, imply a desirable trend of higher level of required {\it compositionality} between text and image (lower modality redundancy). See \Cref{sec:modality_redundancy}.} 
   \label{fig:k-filter}
\end{figure}

\begin{figure}[t]
  \centering
  \includegraphics[width=1\linewidth]{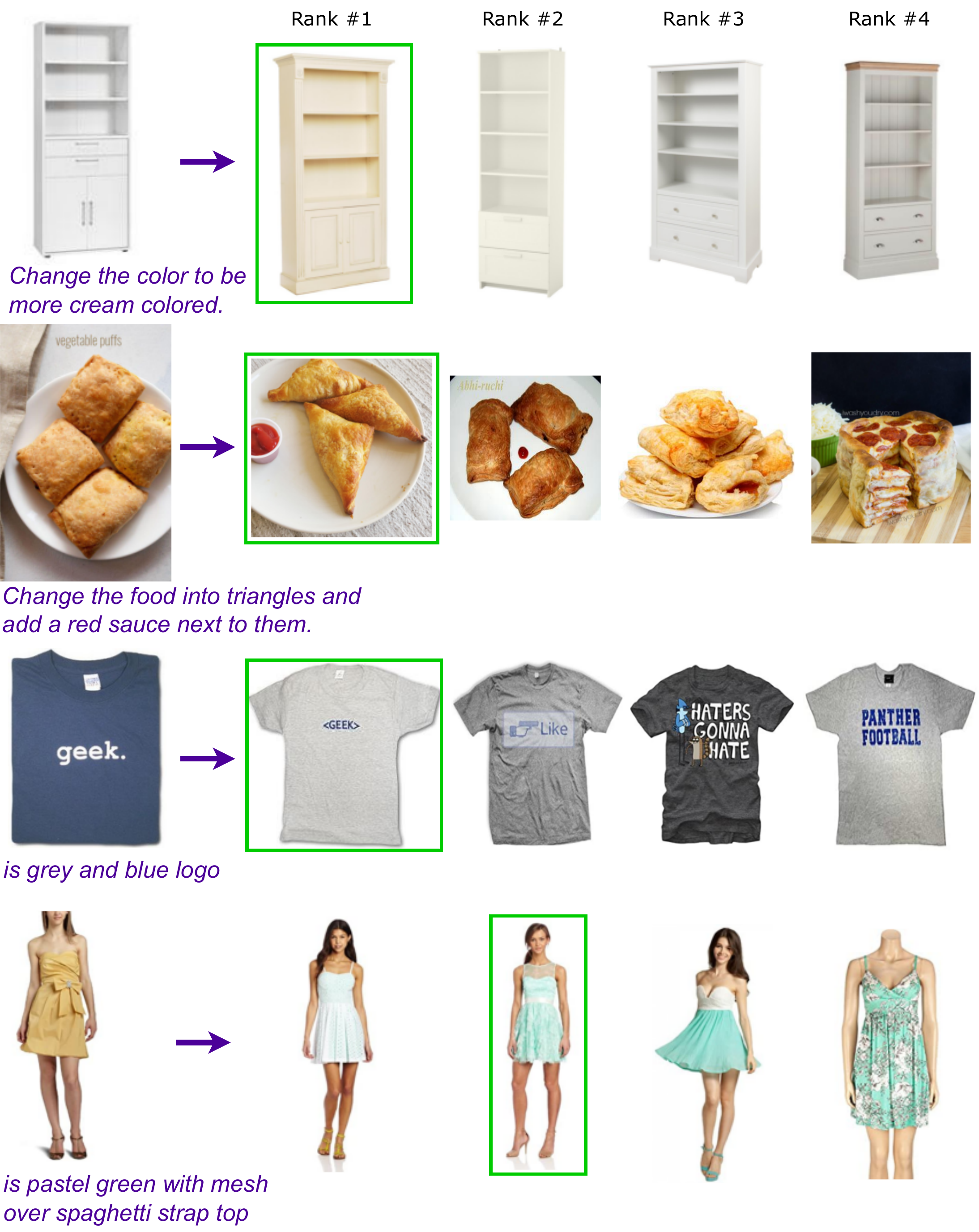}
   \caption{\our top-4 retrievals (from left to right) for queries in \cirr (top two) and \fashioniq (bottom two). The query (image and text) is shown in the left column. The single ground truth target is framed in green. Arguably, additional images could be marked acceptable (referred as false-negatives).}
   \label{fig:fiq_cirr_ex}
\end{figure}

Next, we employ a similar analysis for studying the degree to which \coir methods (trained on a certain dataset) are affected by the presence of modality redundancies in the dataset.
Starting from the full \cirr validation set, denoted as $V$, we generate a sequence of progressively ``purified" subsets $V_n \subset V$, with each subset containing fewer modality redundancies. Specifically, subset $V_n$ is generated by removing from $V$ all of the queries for which the naive CLIP-based Text-to-Image retriever, retrieves the correct target image among it's top-$n$ results. In \Cref{fig:k-filter} we plot the average of Recall$@\{1,5,10,50\}$ as a function of $n$, measured by applying our baseline, \our on each dataset.

Note that the performance of the CLIP-based retriever (blue line) vanishes at $V_{50}$, since by construction, $V_{50}$ contains only queries for which CLIP failed to retrieve the target within its top 50 results. A similar trend is observed with \our-(Text-Only), a variant of our model trained on transition-texts only, ignoring the image. \our-(Text-Only) exhibits performance degradation with increased $n$, as it relies solely on query-text. While \our that was trained on \cirr (orange line) shows better performance, it still suffers some degradation as $n$ grows, implying that some bias towards modality redundancies might still exist. However, when \our is trained or pre-trained on \ourDS dataset (pink and brown lines, respectively), it achieves the best performance, which is roughly constant regardless of $n$.
Thus, \ourDS appears to be effective at removing bias towards redundancies, and \our pre-trained on it is better suited for datasets with high compositionality.

\section{Evaluation}
\label{sec:evaluation}

We evaluate our baseline on \fashioniq, \cirr, and {\it \ourDS} benchmarks, one domain-specific and the others more general and broad, based on natural images. First, we show the results from our newly suggested baseline, CASE. 
Next we examine the effect of using our new \ourDS dataset for training and pre-training (train/val.~split of 92\% \& 8\%). We also present results with pre-training with a mixture of COCO captions, that are very descriptive to better handle samples where the transition text is highly detailed, making the the query image often redundant (\ie text-to-Image retrieval). To this end, we conduct an experiment where we train \our on \ourDS, replacing 50\% of transition-texts $Q_t$, with captions, corresponding to the target image. Namely, we change the train distribution to combine both \coir and text-to-image samples, as discussed in \cref{sec:Data Collection}. We then explain the results thru the properties of different datasets in terms of modality redundancy.
\subsection{Datasets}
\fashioniq \cite{fashioniq} contains crowdsourced descriptions of differences between images of fashion products. Images are collected from the web and divided to three categories of {\it Shirts}, {\it Dresses} and {\it Tops\&Tees}. The query and target images were automatically paired based on title similarities (crawled from the web). This dataset consists of 30K queries (see \Cref{tab:datasets}), annotated on 40.5K different fashion images. There are 4.4K different tokens in the transition-texts (according to {\it BERT} tokenizer). The validation corpus contains 15.4K different images (from which target should be retrieved). The bottom two rows of \Cref{fig:fiq_cirr_ex} show \fashioniq retrieval examples.

\cirr contains open domain natural images, taken from NLVR2 \cite{nlvr2}. It contains a total of 36.5K queries annotated on 19K different images, with 6.8K unique tokens in the transition texts. Examples may be seen in the top two rows of \cref{fig:fiq_cirr_ex}. Its validation corpus is relatively small, with a size of 2.3K. The authors further suggest two benchmarks, one \emph{general}, with the target search space as the entire validation corpus, and a \emph{subset}, where the search space is a subgroup of six images similar to the query image (based on pre-trained ResNet15 feature distance), demonstrating a fine-grained retrieval task. 
\subsection{Implementation Details}
 We set an {\it AdamW} optimizer, initializing learning rate by $5\times 10^{-5}$ with a exponential decay rate of $0.93$ to $1\times 10^{-6}$.
 We train \our on \cirr with a batch size of 2048 for 6 epochs. For \fashioniq, we train with a batch size of 1024 for 20 epochs (further ablation on the batch size is available in suppl. material).
 On \ourDS we train with a batch size of 3840 for 10 epochs. We use the Recall@K surrogate loss~\cite{patel2022recall} as the differentiable version of the Recall@K metric. Training on four A100 nodes takes 0.5-6 minutes per epoch, depending on dataset size.
\subsection{Results}
\label{sec:results}
We start by showing the performance of our \our baseline on {\it FashionIQ}, in \Cref{tab:fiq_val}. The results are broken down to different clothing categories and Recall@K values. For demonstrating modality redundancies, we train two baselines only on query-image (Image-only) or query-text (Text-only).
Interestingly, \our achieves SoTA results, surpassing the previous top performing method (LF-CLIP \cite{cclip}) by a large margin (13.4\% and 11.6\% absolute points at Recall@10,50 respectively). 
The poor results for \our (Image-only) baseline, show that visual information is not sufficient for {\it FashionIQ}, as often the transition text asks for a certain change in the image (see \Cref{sec:modality_redundancy}). However, the \our (Text-only) baseline results are close to the previous method of LF-CLIP, indicating the high level of redundancy in the image, as the single text modality is sufficient to reach the previous SoTA performance. 
\begin{table}[ht]
\begin{center}
\resizebox{\columnwidth}{!}{%
\begin{tabular}{@{}l|llllll|ll@{}}
\toprule
 & \multicolumn{2}{c}{Shirt} & \multicolumn{2}{c}{Dress} & \multicolumn{2}{c|}{Toptee} & \multicolumn{2}{c}{Average} \\
Method & R@10 & R@50 & R@10 & R@50 & R@10 & R@50 & R@10 & R@50 \\ \midrule
Random & 0.16 & 0.79 & 0.26 & 1.31 & 0.19 & 0.95 & 0.06 & 0.32 \\
JVSM {\scriptsize \cite{JVSM}} & 12.0 & 27.1 & 10.7 & 25.9 & 13.0 & 26.9 & 11.9 & 26.6 \\
CIRPLANT {\scriptsize\cite{cirr}}& 17.53 & 38.81 & 17.45 & 40.41 & 61.64 & 45.38 & 18.87 & 41.53 \\
TRACE {\scriptsize \cite{TRACE}} & 20.80 & 40.80 & 22.70 & 44.91 & 24.22 & 49.80 & 22.57 & 46.19 \\ 
VAL  & 22.38 & 44.15 & 22.53 & 44.00 & 27.53 & 51.68 & 24.15 & 46.61 \\ 
MAAF {\scriptsize\cite{MAAF}}& 21.3 & 44.2 & 23.8 & 48.6 & 27.9 & 53.6 & 24.3 & 48.8 \\
CurlingNet {\scriptsize\cite{CurlingNet}}& 21.45 & 44.56 & 26.15 & 53.24 & 30.12 & 55.23 & 25.90 & 51.01 \\
RTIC-GCN {\scriptsize\cite{RTIC}} & 23.79 & 47.25 & 29.15 & 54.04 & 31.61 & 57.98 & 28.18 & 53.09 \\
CoSMo{\scriptsize\cite{Lee_2021_CVPR_cosmo}} & 24.90 & 49.18 & 25.64 & 50.30 & 29.21 & 57.46 & 26.58 & 52.31 \\
ARTEMIS{\scriptsize\cite{ARTEMIS}} & 21.78 & 43.64 & 27.16 & 52.40 & 29.20 & 53.83 & 26.05 & 50.29 \\
DCNet{\scriptsize\cite{Kim_Yu_Kim_Kim_2021_dcnet}} & 23.95 & 47.30 & 28.95 & 56.07 & 30.44 & 58.29 & 27.78 & 53.89 \\
SAC{\scriptsize\cite{SAC}} & 28.02 & 51.86 & 26.52 & 51.01 & 32.70 & 61.23 & 29.08 & 54.70 \\ 
FashionVLP{\scriptsize\cite{FashionVLP}} & 31.89 & \ul{58.44} & 32.42 & \ul{60.29} & \ul{38.51} & \ul{68.79} & 34.27 & \ul{62.51} \\

LF-BLIP & 25.39 & 43.57  &  25.31 & 44.05  &  26.54 & 44.48 & 25.75 & 43.98  \\
LF-CLIP {\scriptsize\cite{cclip}} & \ul{36.36} &  58.00 & 31.63 &  {56.67} &  {38.19} & {62.42} &  \ul{35.39} &  {59.03} \\
\our & \textbf{48.48} & \textbf{70.23} & \textbf{47.44} & \textbf{69.36} & \textbf{50.18} & \textbf{72.24} & \textbf{48.79} & \textbf{70.68} \\
\bottomrule
\our (Image-only) &  6.72 & 17.52 & 8.02 & 18.6 & 7.93 & 18.05 & 7.43 & 17.95\\
\our (Text-only) &  34.84 & 56.93 &  \ul{33.47} & 55.81 & 35.99 & 58.52 & 34.89 & 57.18\\
\bottomrule
\end{tabular}

}
\end{center}
\caption{Recall@K comparison, on {\it FashionIQ} \cite{fashioniq} validation set. Best performance is in bold and second best is underlined. \our outperforms previous results in all metrics by a large margin. CASE is top performing. CASE(Text-only) results indicate high modality redundancy.}
\label{tab:fiq_val}
\end{table}

\Cref{tab:cirr_test} shows results on \cirr. At the top we present the results from previous methods. The right columns $R_{\textit{subset}}@K$, correspond to the fine-grained retrieval task (introduced originally in \cite{cirr}). Here, we show five different variants of our model. As in {\it FashionIQ}, the poor results for \our (Image-Only) imply that the query-image alone is not sufficient for retrieval also on \cirr.
Interestingly, \our(Text-only) reference surpass previous methods in most metrics, further demonstrating the high level of modality redundancy in \cirr as shown in \cref{sec:modality_redundancy}. This baseline is also top performing in the {\it subset} benchmark.  We believe this is caused by the existing image similarity in the subset, making the query image redundant for the task (see visual examples in suppl.~material). 
\begin{table}[t]
\begin{center}
\resizebox{\columnwidth}{!}{%
\begin{tabular}{cl|cccc|ccc}
\hline
\multicolumn{1}{l}{} &  & \multicolumn{4}{c|}{Recall@K} & \multicolumn{3}{c}{$\text{R}_{\text{subset}}$@K} \\
\multicolumn{1}{c}{Mode} & Method & K=1 & K=5 & K=10 & K=50 & K=1 & K=2 & K=3 \\ \hline
\multirow{10}{*}{{\it Train}} & Random & 0.04 & 0.22 & 0.44 & 2.18 & 16.67 & 33.33 & 50.00 \\
& TIRG  & 14.61 & 48.37 & 64.08 & 90.03 & 22.67 & 44.97 & 65.14 \\ 
 & MAAF & 10.31 & 33.03 & 48.30 & 80.06 & 21.05 & 41.81 & 61.60 \\ 
 & MAAF-BERT & 10.12 & 33.10 & 48.01 & 80.57 & 22.04 & 42.41 & 62.14 \\
 & MAAF-RP & 10.22 & 33.32 & 48.68 & 81.84 & 21.41 & 42.17 & 61.60 \\
  & ARTEMIS & 19.96 & 46.10 & 61.31 & 87.73 & 39.99 & 62.20 & 75.67 \\ 
 & CIRPLANT & 19.55 & 52.55 & 68.39 & 92.38 & 39.20 & 63.03 & 79.49 \\ 
 & LF-BLIP & 20.89 & 48.07 & 61.16 & 83.71 & 50.22 & 73.16 & 86.82 \\
 & LF-CLIP & 33.59 & 65.35 & 77.35 & 95.21 & 62.39 & 81.81 & 92.02 \\
 & \our & \textbf{48.00} & \textbf{79.11} & \textbf{87.25} & \textbf{97.57} & \textbf{75.88} & \textbf{90.58} & \textbf{96.00} \\
\cline{1-9}
 \multirow{3}{*}{\begin{tabular}[c]{@{}c@{}}{\it Zero}\\ {\it Shot}\end{tabular}}
& \our Init. & 16.63 & 33.54 & 42.65 & 65.30 & 55.74 & 77.10 & 88.48 \\
 & \our ~- \ourDS & 30.89 & 60.75 & 73.88 & 92.84 & 60.17 & 80.17 & 90.41 \\
 & \our ~- \ourDS.Ca. & \textbf{35.40} & \textbf{65.78} & \textbf{78.53} & \textbf{94.63} & \textbf{64.29} & \textbf{82.66} & \textbf{91.61} \\
 \cline{1-9} 
 \multirow{2}{*}{\begin{tabular}[c]{@{}c@{}}{\it Pre-Train}\end{tabular}}

 & \our Pre-\ourDS & 48.68 & 79.98 & 88.51 & \textbf{97.49} & 76.39 & 90.12 & \textbf{95.86} \\
 & \our Pre-\ourDS.Ca. & \textbf{49.35} & \textbf{80.02} & \textbf{88.75} & 97.47 & \textbf{76.48} & \textbf{90.37} & {95.71} \\ \hline

\cline{1-9} 
 \multirow{2}{*}{\begin{tabular}[c]{@{}c@{}}{\it Modality}\\{\it Redundancy}\end{tabular}}
 & \our (Image-only) & 0.00 & 0.19 & 0.41 & 2.12 & 19.78 & 39.49 & 59.87 \\
 & \our (Text-only) & 39.01 & 69.53 & 79.24 & 91.32 & {78.68} & {91.70} & {96.08} \\
 \hline
\end{tabular}
  }
\end{center}
\caption{Recall@K comparison on \cirr test set. \our shows state-of-the-art results in all cases. 
The methods in the ``Train'' mode were all trained on the \cirr train set, in contrast to ``Zero-shot'' mode. \our (Text/Image-only) was trained solely on a single modality (text/image, respectively), while ignoring the other.
}
\vspace{-1em}
\label{tab:cirr_test}
\end{table}
Next, we observe the performance of \our, which consistently outperforms previous methods. 
We further show visual and textual explainability maps in the supplementary material.

Now we examine the impact of our \ourDS dataset in two main aspects 1) Pr-training and 2) Zero-Shot inference. The results are shown under \our Pre-\ourDS, gaining $\sim$0.5-1\% improvement. We found that this relatively small improvement is due to the modality redundancy in the \cirr dataset. We justify this assumption by the analysis shown in \cref{fig:k-filter}, and by shifting \our towards the distribution of \cirr. Specifically, we train \our on a mix of \ourDS transition-texts with full target captions (taken from COCO captions), which we denote \ourDS.Ca, thus biasing the model towards highly descriptive texts. As shown in \cref{tab:cirr_test}, it further boosts performance on \cirr, a fact that we attribute to improvement of the Text-to-Image (T2I) search capability (see \Cref{sec:modality_redundancy}). 
The use of \ourDS.Ca significantly boosts performance also in \emph{zero-shot} on \cirr test set (\ie without even training on \cirr), surpassing previous methods in most metrics, indicating again the impact of modality redundancy on the results.


Finally, in \Cref{tab:lasco_val}, we benchmark on \ourDS. To this end, we apply the two \our variants (Text-Only, Image-Only) that result in poor performance, implying the necessity of both modalities in this dataset. We further test LF-CLIP \cite{cclip} trained on \ourDS, and observe its significant drop in performance (compared to \cirr), implying that \ourDS dataset introduces a higher true \coir challenge. Finally, \our performs best also here, raising \eg, Recall@1 from 4.01\% (by prior LF-CLIP) to 7.08\%, and Recall@50 from 32.08\% to 50.25\%. 
\begin{table}[ht]
\begin{center}
\resizebox{\columnwidth}{!}{%
\begin{tabular}{lcccccc}
\hline
Method & R@1 & R@5 & R@10 & R@50 & R@500 & R@1000 \\ \hline
Random & 0.00 & 0.01 & 0.03 & 0.13 & 1.26 & 2.51 \\
\our (Image-Only) & 2.21 & 7.39 & 11.82 & 30.62 & 72.64 & 82.46 \\
\our (Text-Only) & 2.39 & 6.89 & 10.39 & 24.92 & 61.23 & 71.23 \\
\hline

LF-CLIP & 4.01 & 10.23 & 14.68 & 32.08 & 72.69 & 83.32 \\
LF-BLIP & 4.26 & 12.01 & 17.11 & 36.54 & 74.62 & 83.55 \\
\our & \textbf{7.08} & \textbf{18.50} & \textbf{26.16} & \textbf{50.25} & \textbf{85.46} & \textbf{91.48} \\
 \hline
\end{tabular}
}
\end{center}
\caption{Results on \ourDS validation set. }
\label{tab:lasco_val}
\end{table}

{\bf Ablation Study:} \label{sec:ablations}
 First, we construct a reference by replacing pre-trained encoders of CLIP, with BLIP's, presented in \Cref{tab:cirr_test,tab:fiq_val,tab:lasco_val}, named LF-BLIP. Next, we ablate various key-components of \our to examine their impact on performance.
\Cref{tab:fiq_ablations} reports ablation results on the frequently used \fashioniq dataset. We observed similar trends also on CIRR (see suppl. material). 
We train \our without the reverse queries objective described in \Cref{sec:method_reverse_queries} (No-RQ). We observe that reverse queries improve performance by $0.5$--$1.2$\% absolute points ($\sim\!\!\!5$\% relative performance boost at R@1). Using surrogate Recall@K loss instead of common contrastive loss, further improves results by roughly $0.5\%$ absolute points. Finally, we examine the influence of fine-tuning our ViT parameters. We observe improvement at higher K values (R@10, R@50) that trades-off with lower R@K metrics (R@1, R@5). However, this experiment is dataset-dependent.
We observed that when training on \ourDS, with 81.6K images, and testing on the \ourDS test set, all metrics were improved by absolute $0.4-2\%$.
\begin{table}[t]
\begin{center}
\resizebox{0.8\columnwidth}{!}{%
\begin{tabular}{@{}llllll@{}}
\toprule
 & R@1 & R@5 & R@10 & R@50 & R@100 \\ \midrule
LF-BLIP &  8.66 & 19.33 & 25.75 & 43.98 & 52.61 \\
No-RQ &  19.98 & 38.36 & 47.57 & {70.28} & 77.68 \\
Contrastive Loss &  20.45 & 38.96 & 48.04 & 69.98 & {78.06} \\
Freeze ViT &  \textbf{21.03} & \textbf{39.43} & 48.25 & 69.85 & 77.96\\
\our &  {20.69} & {39.38} & \textbf{48.79} & \textbf{70.68} & \textbf{78.54}  \\
\bottomrule
\end{tabular}
}
\end{center}
\caption{Ablation study for different configurations of \our, conducted on \fashioniq.}
\label{tab:fiq_ablations}
\vspace{-1em}
\end{table}
\section{Discussion}
We shed more light on the task of \coir. Data labeling for \coir appears to be difficult and costly, exposed to serious biases (\eg, redundancy of the image-query), bounded by inevitable flaws (\eg, false-negatives), and eventually ending up with a low quality and size of data. We suggest a remedy for most of these shortcomings via an inexpensive solution: leveraging labels from a popular related task, to create a new labeled dataset, \ourDS. We extensively analyze current \coir datasets, in order to show their effectiveness, generalization and the capability of a certain model (trained on specific dataset) in handling the desired compositionality in the \coir task. We also suggest the \our baseline, that relies on early fusion of the query modalities through a cross-attention module. We demonstrate the effectiveness of \our by achieving top results on labelled \coir benchmarks from two different domains. To the best of our knowledge, our new baseline, also leverages the smallest dimension of the search space (shared embedding space) among the methods being compared, resulting in a further reduction in computational expenses. 
We believe this work, including our newly introduced dataset, might serve as a useful and practical tool 
not solely limited to the intricate \coir task but also extending to the broader realm of multi-modal learning.

\paragraph{\bf Acknowledgments:} 
This work was supported in part by the Israel Science Foundation (grants 2492/20 and 3611/21). We thank Or Kedar for his assistance in parts of this research. 

{\small
\bibliography{egbib}

\begin{thebibliography}{49}
\providecommand{\natexlab}[1]{#1}

\bibitem[{Agrawal et~al.(2019)Agrawal, Desai, Wang, Chen, Jain, Johnson, Batra,
  Parikh, Lee, and Anderson}]{nocaps2019}
Agrawal, H.; Desai, K.; Wang, Y.; Chen, X.; Jain, R.; Johnson, M.; Batra, D.;
  Parikh, D.; Lee, S.; and Anderson, P. 2019.
\newblock {Nocaps: Novel object captioning at scale}.
\newblock In \emph{{ICCV}}, 8948--8957.

\bibitem[{Antol et~al.(2015)Antol, Agrawal, Lu, Mitchell, Batra, Zitnick, and
  Parikh}]{VQA}
Antol, S.; Agrawal, A.; Lu, J.; Mitchell, M.; Batra, D.; Zitnick, C.~L.; and
  Parikh, D. 2015.
\newblock {VQA}: {V}isual {Q}uestion {A}nswering.
\newblock In \emph{ICCV}.

\bibitem[{Baldrati et~al.(2022)Baldrati, Bertini, Uricchio, and Bimbo}]{cclip}
Baldrati, A.; Bertini, M.; Uricchio, T.; and Bimbo, A.~D. 2022.
\newblock {Effective conditioned and composed image retrieval combining
  CLIP-based features}.
\newblock In \emph{{CVPR}}, 21434--21442. {IEEE}.

\bibitem[{Barz and Denzler(2021)}]{barz2021content}
Barz, B.; and Denzler, J. 2021.
\newblock Content-based image retrieval and the semantic gap in the deep
  learning era.
\newblock In \emph{ICPR}, 245--260.

\bibitem[{Brown et~al.(2020)Brown, Mann, Ryder, Subbiah, Kaplan, Dhariwal,
  Neelakantan, Shyam, Sastry, Askell, Agarwal, Herbert-Voss, Krueger, Henighan,
  Child, Ramesh, Ziegler, Wu, Winter, Hesse, Chen, Sigler, Litwin, Gray, Chess,
  Clark, Berner, McCandlish, Radford, Sutskever, and Amodei}]{GPT_3}
Brown, T.; Mann, B.; Ryder, N.; Subbiah, M.; Kaplan, J.~D.; Dhariwal, P.;
  Neelakantan, A.; Shyam, P.; Sastry, G.; Askell, A.; Agarwal, S.;
  Herbert-Voss, A.; Krueger, G.; Henighan, T.; Child, R.; Ramesh, A.; Ziegler,
  D.; Wu, J.; Winter, C.; Hesse, C.; Chen, M.; Sigler, E.; Litwin, M.; Gray,
  S.; Chess, B.; Clark, J.; Berner, C.; McCandlish, S.; Radford, A.; Sutskever,
  I.; and Amodei, D. 2020.
\newblock {Language Models are Few-Shot Learners}.
\newblock In \emph{{NeurIPS}}, volume~33, 1877--1901.

\bibitem[{Chefer, Gur, and Wolf(2021)}]{Chefer_2021_ICCV}
Chefer, H.; Gur, S.; and Wolf, L. 2021.
\newblock {Generic Attention-Model Explainability for Interpreting Bi-Modal and
  Encoder-Decoder Transformers}.
\newblock In \emph{{ICCV}}, 397--406.

\bibitem[{Chen and Dolan(2011)}]{MSVD}
Chen, D.~L.; and Dolan, W.~B. 2011.
\newblock {Collecting Highly Parallel Data for Paraphrase Evaluation}.
\newblock In Lin, D.; Matsumoto, Y.; and Mihalcea, R., eds., \emph{{ACL}},
  190--200.

\bibitem[{Chen and Bazzani(2020)}]{JVSM}
Chen, Y.; and Bazzani, L. 2020.
\newblock {Learning Joint Visual Semantic Matching Embeddings for
  Language-Guided Retrieval}.
\newblock In \emph{{ECCV}}, volume 12367, 136--152.

\bibitem[{Chen, Gong, and Bazzani(2020)}]{VAL_IR}
Chen, Y.; Gong, S.; and Bazzani, L. 2020.
\newblock {Image Search With Text Feedback by Visiolinguistic Attention
  Learning}.
\newblock In \emph{{CVPR}}, 2998--3008.

\bibitem[{Couairon et~al.(2022)Couairon, Douze, Cord, and
  Schwenk}]{Arithmetic_multimodal_IR}
Couairon, G.; Douze, M.; Cord, M.; and Schwenk, H. 2022.
\newblock {Embedding Arithmetic of Multimodal Queries for Image Retrieval}.
\newblock In \emph{{CVPRW}}, 4946--4954. {IEEE}.

\bibitem[{Das et~al.(2017)Das, Kottur, Gupta, Singh, Yadav, Moura, Parikh, and
  Batra}]{visdial}
Das, A.; Kottur, S.; Gupta, K.; Singh, A.; Yadav, D.; Moura, J.~M.; Parikh, D.;
  and Batra, D. 2017.
\newblock {V}isual {D}ialog.
\newblock In \emph{CVPR}.

\bibitem[{Delmas et~al.(2022)Delmas, de~Rezende, Csurka, and Larlus}]{ARTEMIS}
Delmas, G.; de~Rezende, R.~S.; Csurka, G.; and Larlus, D. 2022.
\newblock {ARTEMIS: Attention-based Retrieval with Text-Explicit Matching and
  Implicit Similarity}.
\newblock In \emph{The Tenth International Conference on Learning
  Representations, {ICLR} 2022, Virtual Event, April 25-29, 2022}.
  OpenReview.net.

\bibitem[{Devlin et~al.(2019)Devlin, Chang, Lee, and
  Toutanova}]{devlin2019bert}
Devlin, J.; Chang, M.; Lee, K.; and Toutanova, K. 2019.
\newblock {BERT: Pre-training of Deep Bidirectional Transformers for Language
  Understanding}.
\newblock In \emph{{NAACL-HLT}}.

\bibitem[{Dodds et~al.(2020)Dodds, Culpepper, Herdade, Zhang, and
  Boakye}]{MAAF}
Dodds, E.; Culpepper, J.; Herdade, S.; Zhang, Y.; and Boakye, K. 2020.
\newblock {Modality-Agnostic Attention Fusion for visual search with text
  feedback}.
\newblock \emph{CoRR}, abs/2007.00145.

\bibitem[{Dosovitskiy et~al.(2021)Dosovitskiy, Beyer, Kolesnikov, Weissenborn,
  Zhai, Unterthiner, Dehghani, Minderer, Heigold, Gelly, Uszkoreit, and
  Houlsby}]{DosovitskiyB0WZ21}
Dosovitskiy, A.; Beyer, L.; Kolesnikov, A.; Weissenborn, D.; Zhai, X.;
  Unterthiner, T.; Dehghani, M.; Minderer, M.; Heigold, G.; Gelly, S.;
  Uszkoreit, J.; and Houlsby, N. 2021.
\newblock {An Image is Worth 16x16 Words: Transformers for Image Recognition at
  Scale}.
\newblock In \emph{{ICLR}}.

\bibitem[{Dubey(2021)}]{dubey2021decade}
Dubey, S.~R. 2021.
\newblock A decade survey of content based image retrieval using deep learning.
\newblock \emph{IEEE Transactions on Circuits and Systems for Video
  Technology}, 32(5): 2687--2704.

\bibitem[{Goenka et~al.(2022)Goenka, Zheng, Jaiswal, Chada, Wu, Hedau, and
  Natarajan}]{FashionVLP}
Goenka, S.; Zheng, Z.; Jaiswal, A.; Chada, R.; Wu, Y.; Hedau, V.; and
  Natarajan, P. 2022.
\newblock {FashionVLP: Vision Language Transformer for Fashion Retrieval with
  Feedback}.
\newblock In \emph{{CVPR}}, 14085--14095. {IEEE}.

\bibitem[{Goyal et~al.(2017)Goyal, Khot, Summers{-}Stay, Batra, and
  Parikh}]{balanced_vqa_v2}
Goyal, Y.; Khot, T.; Summers{-}Stay, D.; Batra, D.; and Parikh, D. 2017.
\newblock {Making the {V} in {VQA} Matter: Elevating the Role of Image
  Understanding in Visual Question Answering}.
\newblock In \emph{{CVPR}}.

\bibitem[{Guo et~al.(2018)Guo, Wu, Cheng, Rennie, Tesauro, and
  Feris}]{DialogBasedIIR}
Guo, X.; Wu, H.; Cheng, Y.; Rennie, S.; Tesauro, G.; and Feris, R.~S. 2018.
\newblock {Dialog-based Interactive Image Retrieval}.
\newblock In \emph{{NeurIPS}}, 676--686.

\bibitem[{Han et~al.(2017)Han, Wu, Huang, Zhang, Zhu, Li, Zhao, and
  Davis}]{fashion200k}
Han, X.; Wu, Z.; Huang, P.~X.; Zhang, X.; Zhu, M.; Li, Y.; Zhao, Y.; and Davis,
  L.~S. 2017.
\newblock {Automatic Spatially-Aware Fashion Concept Discovery}.
\newblock In \emph{{ICCV}}, 1472--1480.

\bibitem[{He et~al.(2016)He, Zhang, Ren, and Sun}]{resnet}
He, K.; Zhang, X.; Ren, S.; and Sun, J. 2016.
\newblock {Deep Residual Learning for Image Recognition}.
\newblock In \emph{{CVPR}}.

\bibitem[{Hosseinzadeh and Wang(2020)}]{HosseinzadehW20}
Hosseinzadeh, M.; and Wang, Y. 2020.
\newblock {Composed Query Image Retrieval Using Locally Bounded Features}.
\newblock In \emph{{CVPR}}, 3593--3602. Computer Vision Foundation / {IEEE}.

\bibitem[{Isola, Lim, and Adelson(2015)}]{MITstates}
Isola, P.; Lim, J.~J.; and Adelson, E.~H. 2015.
\newblock {Discovering states and transformations in image collections}.
\newblock In \emph{{CVPR}}, 1383--1391.

\bibitem[{Jandial et~al.(2022)Jandial, Badjatiya, Chawla, Chopra, Sarkar, and
  Krishnamurthy}]{SAC}
Jandial, S.; Badjatiya, P.; Chawla, P.; Chopra, A.; Sarkar, M.; and
  Krishnamurthy, B. 2022.
\newblock {SAC: Semantic Attention Composition for Text-Conditioned Image
  Retrieval}.
\newblock In \emph{{WACV}}, 4021--4030.

\bibitem[{Jandial et~al.(2020)Jandial, Chopra, Badjatiya, Chawla, Sarkar, and
  Krishnamurthy}]{TRACE}
Jandial, S.; Chopra, A.; Badjatiya, P.; Chawla, P.; Sarkar, M.; and
  Krishnamurthy, B. 2020.
\newblock {TRACE: Transform Aggregate and Compose Visiolinguistic
  Representations for Image Search with Text Feedback}.
\newblock \emph{CoRR}, abs/2009.01485.

\bibitem[{Kim et~al.(2021)Kim, Yu, Kim, and Kim}]{Kim_Yu_Kim_Kim_2021_dcnet}
Kim, J.; Yu, Y.; Kim, H.; and Kim, G. 2021.
\newblock {Dual Compositional Learning in Interactive Image Retrieval}.
\newblock \emph{{AAAI}}, 35(2): 1771--1779.

\bibitem[{Lee, Kim, and Han(2021)}]{Lee_2021_CVPR_cosmo}
Lee, S.; Kim, D.; and Han, B. 2021.
\newblock {CoSMo: Content-Style Modulation for Image Retrieval With Text
  Feedback}.
\newblock In \emph{{CVPR}}, 802--812.

\bibitem[{Levy, Ben-Ari, and Lischinski(2022)}]{levy2022classification}
Levy, M.; Ben-Ari, R.; and Lischinski, D. 2022.
\newblock {Classification-Regression for Chart comprehension}.
\newblock In \emph{{ECCV}}, 469--484.

\bibitem[{Li et~al.(2022)Li, Li, Xiong, and Hoi}]{BLIP}
Li, J.; Li, D.; Xiong, C.; and Hoi, S. C.~H. 2022.
\newblock {BLIP: Bootstrapping Language-Image Pre-training for Unified
  Vision-Language Understanding and Generation}.
\newblock In \emph{{ICML}}, 12888--12900.

\bibitem[{Li et~al.(2021)Li, Selvaraju, Gotmare, Joty, Xiong, and
  Hoi}]{Align_before_use}
Li, J.; Selvaraju, R.~R.; Gotmare, A.; Joty, S.~R.; Xiong, C.; and Hoi, S.~C.
  2021.
\newblock {Align before Fuse: Vision and Language Representation Learning with
  Momentum Distillation}.
\newblock In Ranzato, M.; Beygelzimer, A.; Dauphin, Y.~N.; Liang, P.; and
  Vaughan, J.~W., eds., \emph{{NeurIPS}}, 9694--9705.

\bibitem[{Li et~al.(2020)Li, Yin, Li, Zhang, Hu, Zhang, Wang, Hu, Dong, Wei,
  Choi, and Gao}]{li2020oscar}
Li, X.; Yin, X.; Li, C.; Zhang, P.; Hu, X.; Zhang, L.; Wang, L.; Hu, H.; Dong,
  L.; Wei, F.; Choi, Y.; and Gao, J. 2020.
\newblock Oscar: Object-Semantics Aligned Pre-training for Vision-Language
  Tasks.
\newblock In \emph{{ECCV}}.

\bibitem[{Lin et~al.(2014)Lin, Maire, Belongie, Hays, Perona, Ramanan,
  Doll{\'{a}}r, and Zitnick}]{coco}
Lin, T.; Maire, M.; Belongie, S.~J.; Hays, J.; Perona, P.; Ramanan, D.;
  Doll{\'{a}}r, P.; and Zitnick, C.~L. 2014.
\newblock {Microsoft COCO: Common Objects in Context}.
\newblock In \emph{{ECCV}}, volume 8693, 740--755.

\bibitem[{Liu et~al.(2021)Liu, Rodriguez-Opazo, Teney, and Gould}]{cirr}
Liu, Z.; Rodriguez-Opazo, C.; Teney, D.; and Gould, S. 2021.
\newblock {Image Retrieval on Real-life Images with Pre-trained
  Vision-and-Language Models}.
\newblock In \emph{{ICCV}}, 2105--2114.

\bibitem[{Lu et~al.(2019)Lu, Batra, Parikh, and Lee}]{vilbert}
Lu, J.; Batra, D.; Parikh, D.; and Lee, S. 2019.
\newblock {ViLBERT: Pretraining Task-Agnostic Visiolinguistic Representations
  for Vision-and-Language Tasks}.
\newblock In \emph{{NeurIPS}}, 13--23.

\bibitem[{Miech et~al.(2019)Miech, Zhukov, Alayrac, Tapaswi, Laptev, and
  Sivic}]{howto100M}
Miech, A.; Zhukov, D.; Alayrac, J.-B.; Tapaswi, M.; Laptev, I.; and Sivic, J.
  2019.
\newblock Howto100m: Learning a text-video embedding by watching hundred
  million narrated video clips.
\newblock In \emph{CVPR}, 2630--2640.

\bibitem[{Nagrani et~al.(2022)Nagrani, Seo, Seybold, Hauth, Manen, Sun, and
  Schmid}]{nagrani2022learning}
Nagrani, A.; Seo, P.~H.; Seybold, B.; Hauth, A.; Manen, S.; Sun, C.; and
  Schmid, C. 2022.
\newblock Learning Audio-Video Modalities from Image Captions.
\newblock \emph{ECCV}.

\bibitem[{Patel, Tolias, and Matas(2022)}]{patel2022recall}
Patel, Y.; Tolias, G.; and Matas, J. 2022.
\newblock Recall@k surrogate loss with large batches and similarity mixup.
\newblock In \emph{{CVPR}}, 7502--7511.

\bibitem[{Radford et~al.(2021)Radford, Kim, Hallacy, Ramesh, Goh, Agarwal,
  Sastry, Askell, Mishkin, Clark, Krueger, and Sutskever}]{clip}
Radford, A.; Kim, J.~W.; Hallacy, C.; Ramesh, A.; Goh, G.; Agarwal, S.; Sastry,
  G.; Askell, A.; Mishkin, P.; Clark, J.; Krueger, G.; and Sutskever, I. 2021.
\newblock {Learning Transferable Visual Models From Natural Language
  Supervision}.
\newblock In Meila, M.; and Zhang, T., eds., \emph{{ICML}}.

\bibitem[{Shin et~al.(2021)Shin, Cho, Ko, and Gu}]{RTIC}
Shin, M.; Cho, Y.; Ko, B.; and Gu, G. 2021.
\newblock {RTIC: Residual Learning for Text and Image Composition using Graph
  Convolutional Network}.
\newblock \emph{CoRR}, abs/2104.03015.

\bibitem[{Suhr et~al.(2019)Suhr, Zhou, Zhang, Zhang, Bai, and Artzi}]{nlvr2}
Suhr, A.; Zhou, S.; Zhang, A.; Zhang, I.; Bai, H.; and Artzi, Y. 2019.
\newblock {A Corpus for Reasoning about Natural Language Grounded in
  Photographs}.
\newblock In \emph{{ACL}}, 6418--6428.

\bibitem[{Vasu et~al.(2021)Vasu, Hu, Dong, Collins, and
  Hoogs}]{vasu2021explainableCBIR}
Vasu, B.; Hu, B.; Dong, B.; Collins, R.; and Hoogs, A. 2021.
\newblock {Explainable, interactive c ontent-based image retrieval}.
\newblock \emph{Applied AI Letters}, 2(4): e41.

\bibitem[{Vaswani et~al.(2017)Vaswani, Shazeer, Parmar, Uszkoreit, Jones,
  Gomez, Kaiser, and Polosukhin}]{vaswani2017attention}
Vaswani, A.; Shazeer, N.; Parmar, N.; Uszkoreit, J.; Jones, L.; Gomez, A.~N.;
  Kaiser, {\L}.; and Polosukhin, I. 2017.
\newblock {Attention Is All You Need}.
\newblock In \emph{NeurIPS}.

\bibitem[{Vo et~al.(2019)Vo, Jiang, Sun, Murphy, Li, Fei-Fei, and Hays}]{tirg}
Vo, N.; Jiang, L.; Sun, C.; Murphy, K.; Li, L.-J.; Fei-Fei, L.; and Hays, J.
  2019.
\newblock {Composing Text and Image for Image Retrieval - an Empirical
  Odyssey}.
\newblock In \emph{{CVPR}}, 6432--6441.

\bibitem[{Wu et~al.(2021)Wu, Gao, Guo, Al{-}Halah, Rennie, Grauman, and
  Feris}]{fashioniq}
Wu, H.; Gao, Y.; Guo, X.; Al{-}Halah, Z.; Rennie, S.; Grauman, K.; and Feris,
  R. 2021.
\newblock {Fashion IQ: A New Dataset Towards Retrieving Images by Natural
  Language Feedback}.
\newblock In \emph{{CVPR}}, 11307--11317.

\bibitem[{Xu et~al.(2016)Xu, Mei, Yao, and Rui}]{MSR_VTT}
Xu, J.; Mei, T.; Yao, T.; and Rui, Y. 2016.
\newblock {MSR-VTT: A Large Video Description Dataset for Bridging Video and
  Language}.
\newblock In \emph{{CVPR}}, 5288--5296. {IEEE} Computer Society.

\bibitem[{Yang et~al.(2021)Yang, Miech, Sivic, Laptev, and
  Schmid}]{justAsk_VQA_2021}
Yang, A.; Miech, A.; Sivic, J.; Laptev, I.; and Schmid, C. 2021.
\newblock {Just ask: Learning to answer questions from millions of narrated
  videos}.
\newblock In \emph{{ICCV}}, 1686--1697.

\bibitem[{Young et~al.(2014)Young, Lai, Hodosh, and Hockenmaier}]{flickr}
Young, P.; Lai, A.; Hodosh, M.; and Hockenmaier, J. 2014.
\newblock {From image descriptions to visual denotations: New similarity
  metrics for semantic inference over event descriptions}.
\newblock \emph{{Trans. Assoc. Comput. Linguistics}}, 2: 67--78.

\bibitem[{Yu et~al.(2020)Yu, Lee, Choi, and Kim}]{CurlingNet}
Yu, Y.; Lee, S.; Choi, Y.; and Kim, G. 2020.
\newblock {CurlingNet: Compositional Learning between Images and Text for
  Fashion {IQ} Data}.
\newblock \emph{CoRR}, abs/2003.12299.

\bibitem[{Zhong, Chen, and Qian(2020)}]{fewshotIR_ICIP2020}
Zhong, Q.; Chen, L.; and Qian, Y. 2020.
\newblock {Few-shot learning for remote sensing image retrieval with maml}.
\newblock In \emph{{ICIP}}, 2446--2450. IEEE.

\end{thebibliography}
}

\newpage
\appendix
\section*{\LARGE{Supplementary Material}}

In this supplementary material we start in \Cref{sec:models_background} by providing more background about previous \coir methods. In \cref{sec:case_fusion} we elaborate on the cross-attention module in our baseline called CASE (Cross-Attention driven Shift Encoder). Then we describe our Text and Image only variants, \ie, CASE (Image-only) and CASE(Text-only), shown in the Evaluation section of the paper (Sec.~6). \Cref{sec:retrieval_exaples} includes additional retrieval examples from our test bed namely {\it FashionIQ}~\cite{fashioniq}, \cirr~\cite{cirr}, as well as our newly suggested dataset LaSCo. 
Next, in \cref{sec:lasco_creation} we show more examples on creation of our transition texts for the triples and our user-study quality assessment. \cref{sec:modality_redundancy_supp} points out more examples on the {\it Modality Redundancy}. Additional ablation analysis of the suggested {\it CASE} architecture are shown in \cref{sec:ablation_study} We conclude the supplementary material in \cref{sec:explainability}, by an explainability analysis showing visualizations over text and image. We intend to release the dataset and make it publicly available, upon acceptance.

\section{Background for CASE baseline}
\label{sec:models_background}
To keep this paper self-contained, we elaborate here on some recent CoIR methods. This also serves to offer insights into our CASE baseline and it's capabilities.

Various architectures has been proposed for CoIR task. \citet{VAL_IR} suggested VAL ({\it Visiolinguistic Attention Learning}) that uses LSTM for query text encoding and CNN for image encoding. They further conduct hierarchical fusion at different CNN layers with objective to match feature maps of query image, conditioned on query text, to the target image.  \citet{HosseinzadehW20} proposed a dot-product attention mechanism for fusing textual features with image region features, based on a region proposal network. Regional and textual features are separately extracted and merged by early fusion. However, \citet{ARTEMIS} reported a better performance also using two different experts, but in a late-fusion manner, fusing global features of the image and text.

\citet{cirr}, in addition to introducing the new \cirr dataset, suggest a new scheme dubbed CIRPLANT. Their method leverages a rich pretrained V\&L model for modifying visual features conditioned on natural language, so as to match those of the target image. Their architecture consists of a ResNet \cite{resnet} feature extractor and a projection FC-layer. A transformer is further used for text encoding and V\&L fusion. 

While utilization of foundation V\&L encoders such as CLIP showed a performance boost in \coir, they are still used in a ``late fusion'' mode, where image and text are separately encoded (each as a single global vector). Other works take a middle way by letting text tokens to interact with a single global visual feature, or with a few selected region-based image features, bounding boxes, \etc. Leveraging the strong capabilities of the most recent VLM foundation models (\eg BLIP) we build a new baseline that combines early fusion with a strong VLM. 
The early fusion in CASE, utilizes a local cross-attention module. It allows textual-visual interactions at token-patch level through multiple intermediate cross-attention layers. In training, we add a new auxiliary task and further replace the common contrastive loss used in previous \coir methods with a Recall@K surrogate loss \cite{patel2022recall} that matches the evaluation measure.

\section{\our Modality Fusion}
\label{sec:case_fusion}
In this section we elaborate on the cross-attention module introduced in Sec.~4 of the paper and \our single modality baselines.

Let us denote $attn_d$ as the original self-attention layer \cite{vaswani2017attention}, applied on vectors in dimension $d\in \mathbb{N}$. Denote the Query, Key and Value of the text modality (inserted into our {\it Shift Encoder} module), obtained from a self-attention layer at certain block, as $Q_t, K_t, V_t \in \mathbb{R}^{n_t\times d}$ where $n_t$ denotes the number of the text tokens and $d$ the representation dimension. We denote by $Q_v, K_v, V_v \in \mathbb{R}^{n_v\times d}$ the Query, Key and Value obtained from the Visual Transformer (ViT) output, with $n_v$ as the number of visual tokens. Bi-modal fusion is then obtained by inter-changing the self-attention matrices between the modalities, as the following:
\begin{eqnarray}
    S = attn_d(Q_t, K_v, V_v) := \mathit{softmax}(\frac{Q_t {K_v}^T}{\sqrt{d}}) V_v \in \mathbb{R}^{n_t\times d}
\end{eqnarray}
Note that $Q_t, Q_v$ are exchanged, to allow information flow or fusion between the modalities. The output $S$ is then fed to a Feed-Forward layer (see Fig.~1 in the paper). 

{\bf Single Modality CASE:} In order to evaluate the impact of each modality separately on CASE, we present in the paper two baselines trained on a single modality: Text-Only and Image-Only. For the first, we set the input image to a tensor of zeros, resulting in a fixed ``image'' input. For the latter, we mask the text by inserting a constant token of [CLS], excluding any other textual information.

\begin{figure*}[t]
	\centering
	\includegraphics[width=0.9\linewidth]{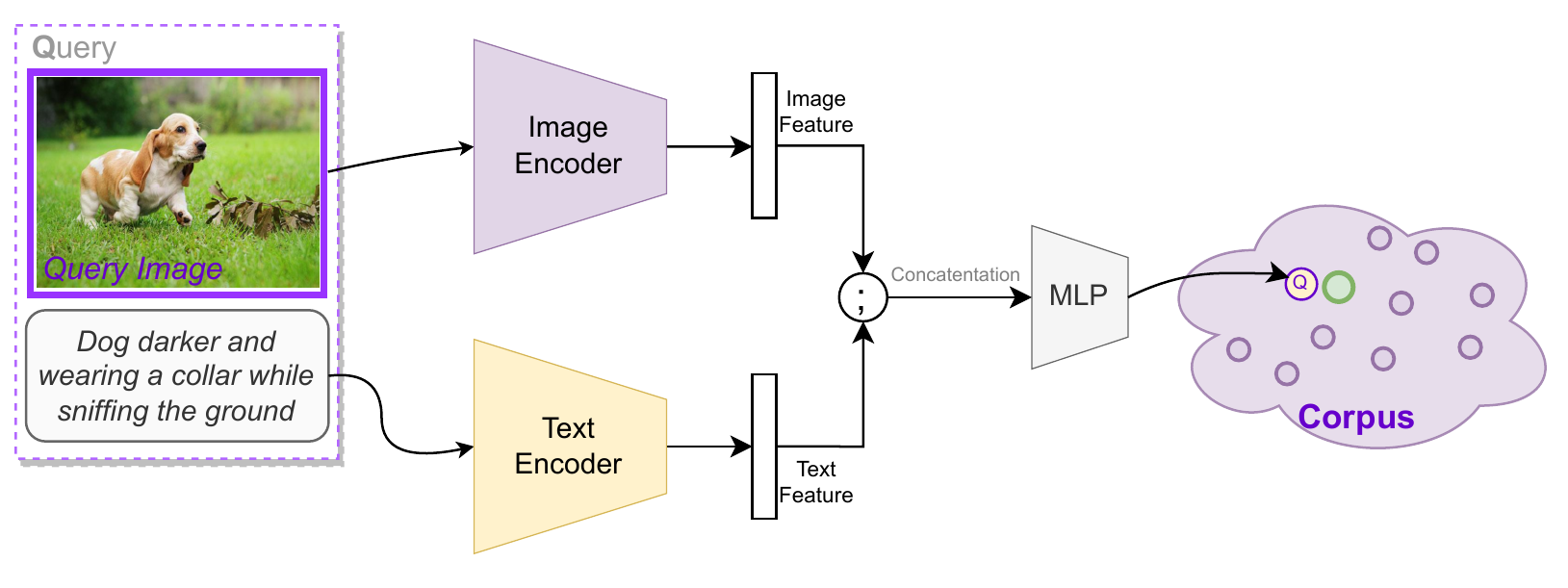}
	\caption{An overview of Late Fusion (LF) architecture. First, each modality is separately encoded with its proper expert (encoder) to a single global feature vector. Next, these two global features are concatenated and being fed to a learnable MLP, with a final output of a single vector. When using CLIP components (with associated pre-trained weights) this architecture introduces the last CoIR SoTA method of \cite{cclip}.}
	\label{fig:lf_arc}
\end{figure*}

{\bf Late Fusion (LF) architecture:}
The paper mentions and describes the LF baselines, referred to as LF-CLIP or LF-BLIP. \Cref{fig:lf_arc} shows an overview of the LF architecture. The weights of the image/text encoders are initialized alternatively with pre-trained weights of CLIP \cite{clip} or BLIP \cite{BLIP}, for LF-CLIP or LF-BLIP, respectively.

\section{Retrieval Examples}
\label{sec:retrieval_exaples}
Here we show retrieval examples of \our, for different validation queries on several datasets.
Note that in evaluation, every query is labeled with only one target image, ignoring answers that might be argued as acceptable (false-negatives). Inevitably, this phenomenon appears in larger image corpora.

\begin{figure*}[t]
  \centering
  \includegraphics[width=1\textwidth,height=0.95\textheight]{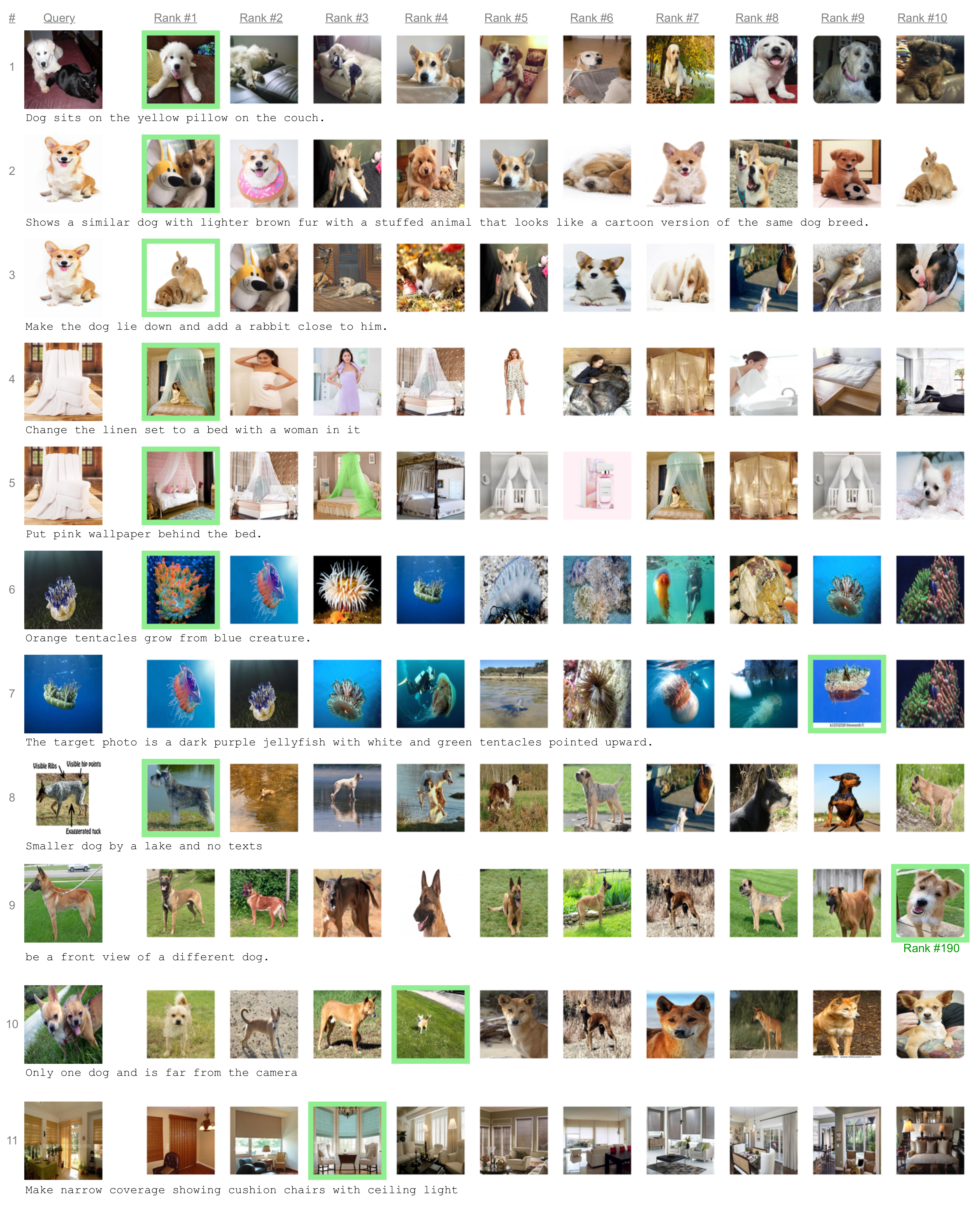}
   \caption{\our retrievals on \cirr validation set. For convenience, images were resized to the same shape. Labeled target image (ground truth) is framed in green. Rows 1-6 demonstrate {\it Modality Redundancy} as they were also successfully retrieved based only on the given transition text (without using the query image).}
   \label{fig:cirr_retrievals}
\end{figure*}

{\bf CIRR:} \Cref{fig:cirr_retrievals} shows the performance of \our on \cirr examples. There are arguably more than a single acceptable result in several cases. Sometimes, the reason behind tagging the specific target image is unclear due to existence of several similar images. This can be an outcome of the CIRR dataset creation protocol, where the annotator is naturally not exposed to the whole dataset, a fact that further implies the challenge associated with creating a ``clean" CoIR large corpus. This leads to an increase in the count of ``false-negatives" (due declination of the labeled target in the rank). See for instance, rows 6, 8, and 9. Another shortcoming observable in these examples (\Cref{fig:cirr_retrievals}) is the {\it Modality-Redundancy} (see \Cref{sec:modality_redundancy_supp}).

\begin{figure*}[t]
  \centering
  \includegraphics[width=1\textwidth,height=0.91\textheight]{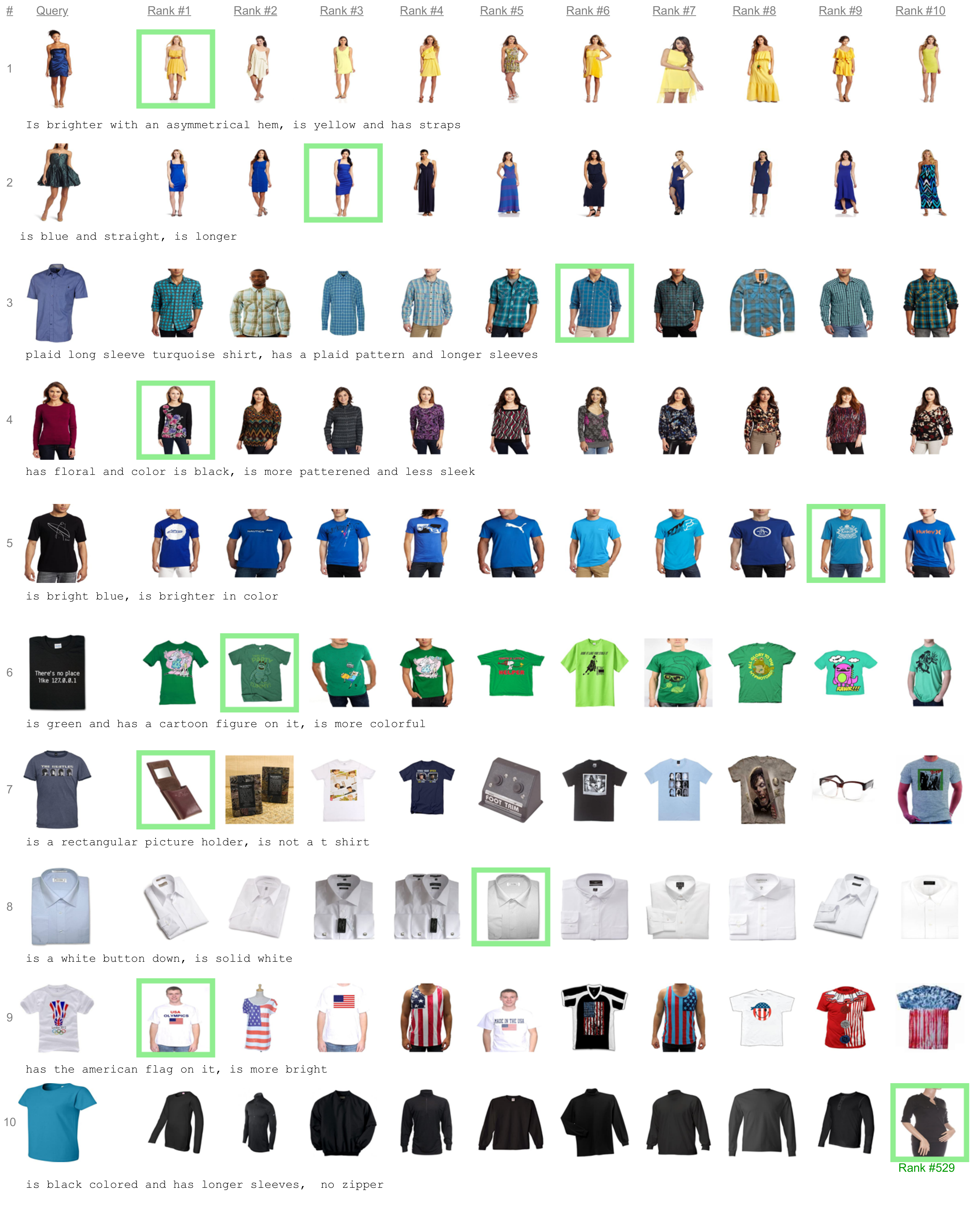}
   \caption{\our retrievals on \fashioniq validation set. For convenience, images were resized to the same shape. Labeled target image (ground truth) is framed in green. The example in the last row, indicates a case where the corpus contains many images similar to the target and can be acceptable retrievals. The true target was ranked 529 here.}
   \label{fig:fiq_retrievals}
\end{figure*}

{\bf FashionIQ:} Several retrieval examples are shown in \Cref{fig:fiq_retrievals}. In numerous instances, the top 5-10 retrievals are very similar and can be arguably considered as legitimate answers.

\begin{figure*}[t]
  \centering
  \includegraphics[width=1\textwidth]{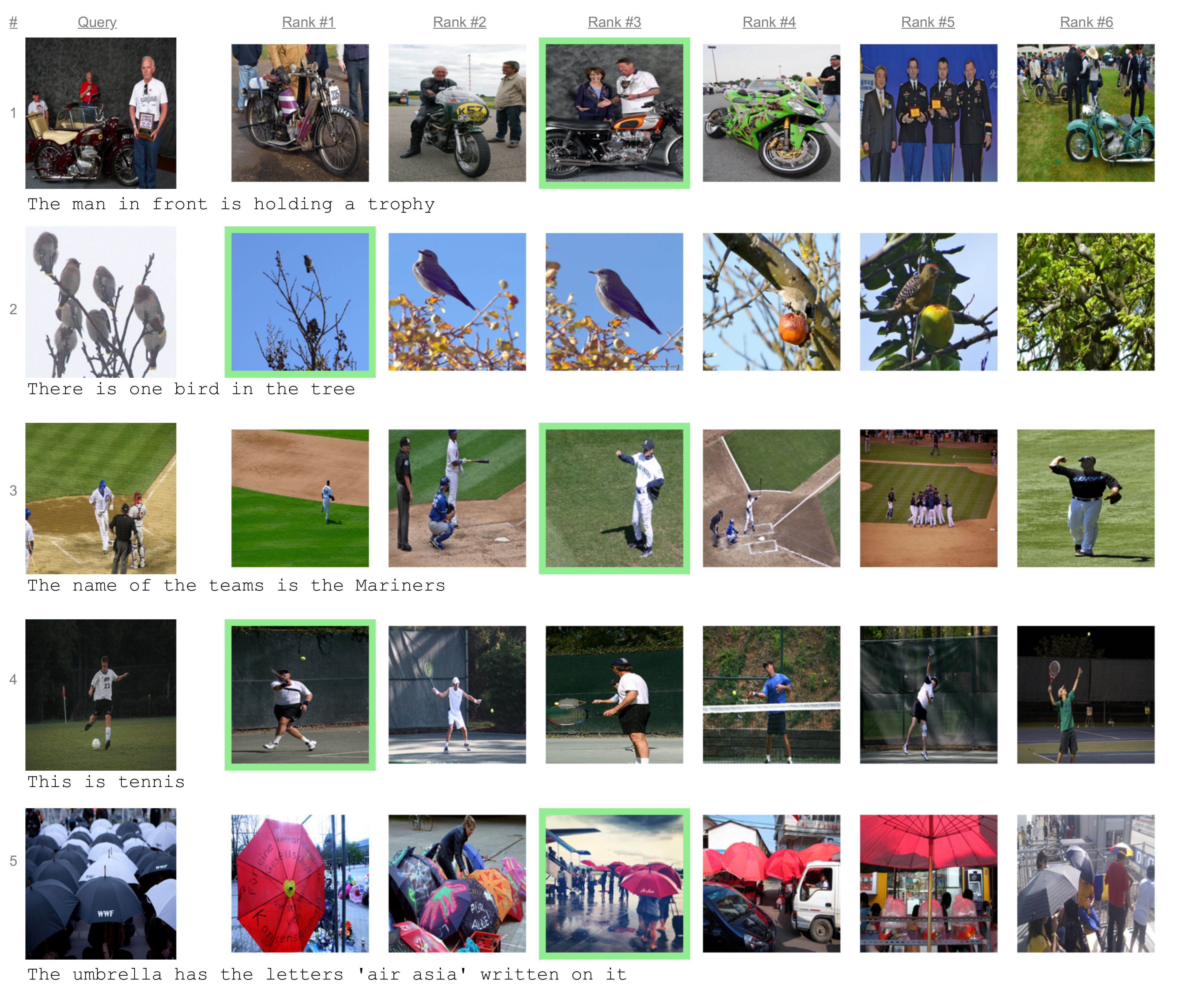}
   \caption{\our retrievals on \ourDS validation set. For convenience, images were resized to the same shape. Labeled target image (ground truth) is framed in green.}
   \label{fig:lasco_retrievals_1}
\end{figure*}

\begin{figure*}[t]
  \centering
  \includegraphics[width=1\textwidth]{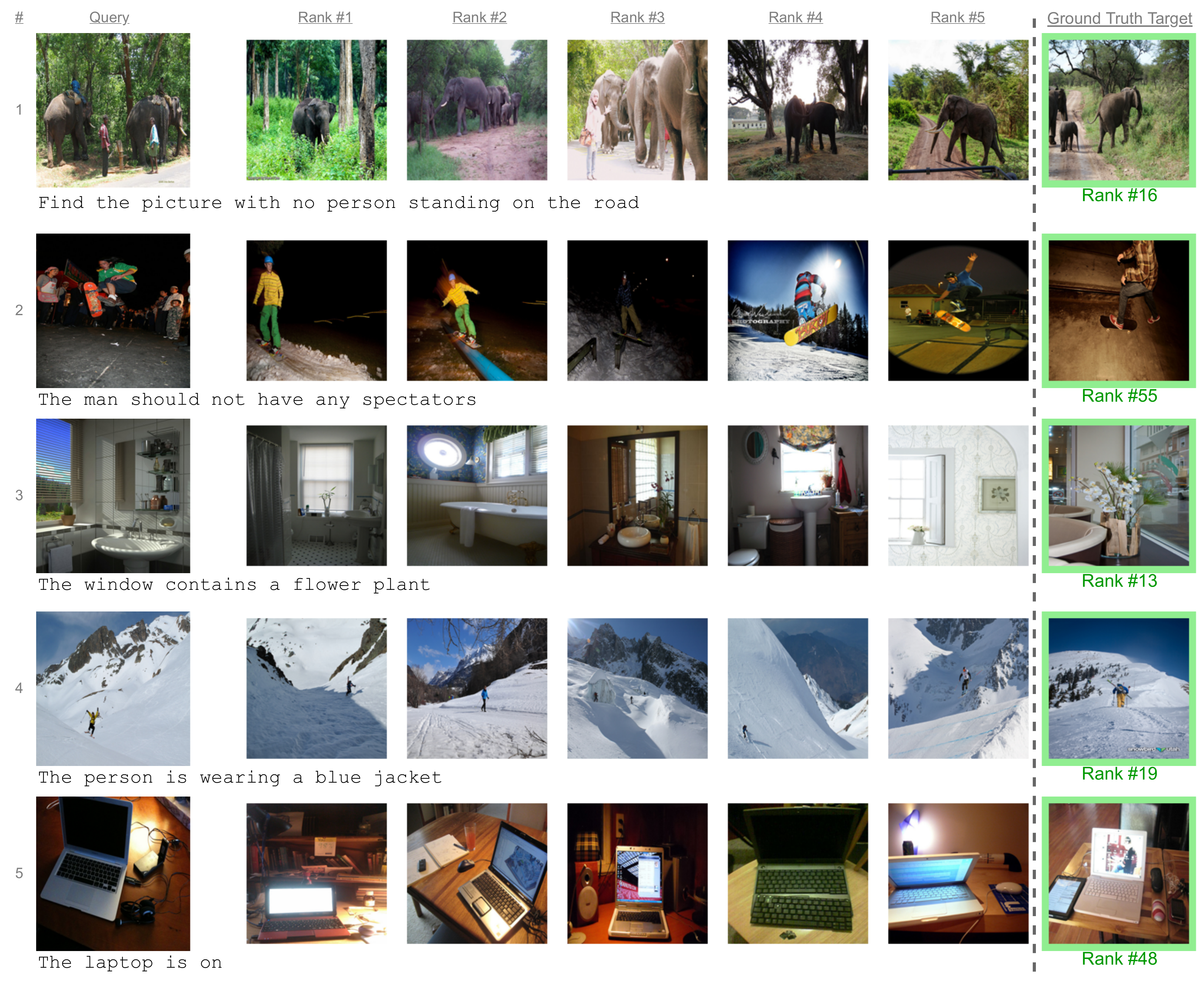}
   \caption{\our retrievals on \ourDS validation set. Examples where the target was not ranked at top-5. The right column shows the ground-truth target, with the associated CASE rank underneath. Note however, although the target was missed, the top-5 images are often acceptable answers. For convenience, images were resized to the same shape. Labeled target image (ground truth) is framed in green.}
   \label{fig:lasco_retrievals_2}
\end{figure*}

{\bf LaSCo:} \Cref{fig:lasco_retrievals_1,fig:lasco_retrievals_2} show several retrieval examples on \ourDS validation set. The results show that our \our model can handle broad concepts from an open domain, having a rich vocabulary. An interesting example is further shown in the last row in \Cref{fig:lasco_retrievals_1}, implying that the model attempts to take into consideration text that appears in an image, 
a consequence of learning from the rich \ourDS dataset. Note that \ourDS validation set includes $39.8$K images, a few order of magnitudes larger than previous datasets, which causes it to inevitably contain false-negatives (See \Cref{fig:lasco_retrievals_2}).


\section{LaSCo Dataset Creation}
\label{sec:lasco_creation}
In this section we show more examples of \ourDS samples.
As described in Sec.~3 in the main paper, query/target images are paired based on samples and their ``complementary'' images in VQA2.0. To create the transition-texts we exploit the GPT-3 few-shot \cite{GPT_3} inference capabilities, in the following way.

\begin{figure*}[t]
  \centering
  \includegraphics[width=1\textwidth,height=0.95\textheight]{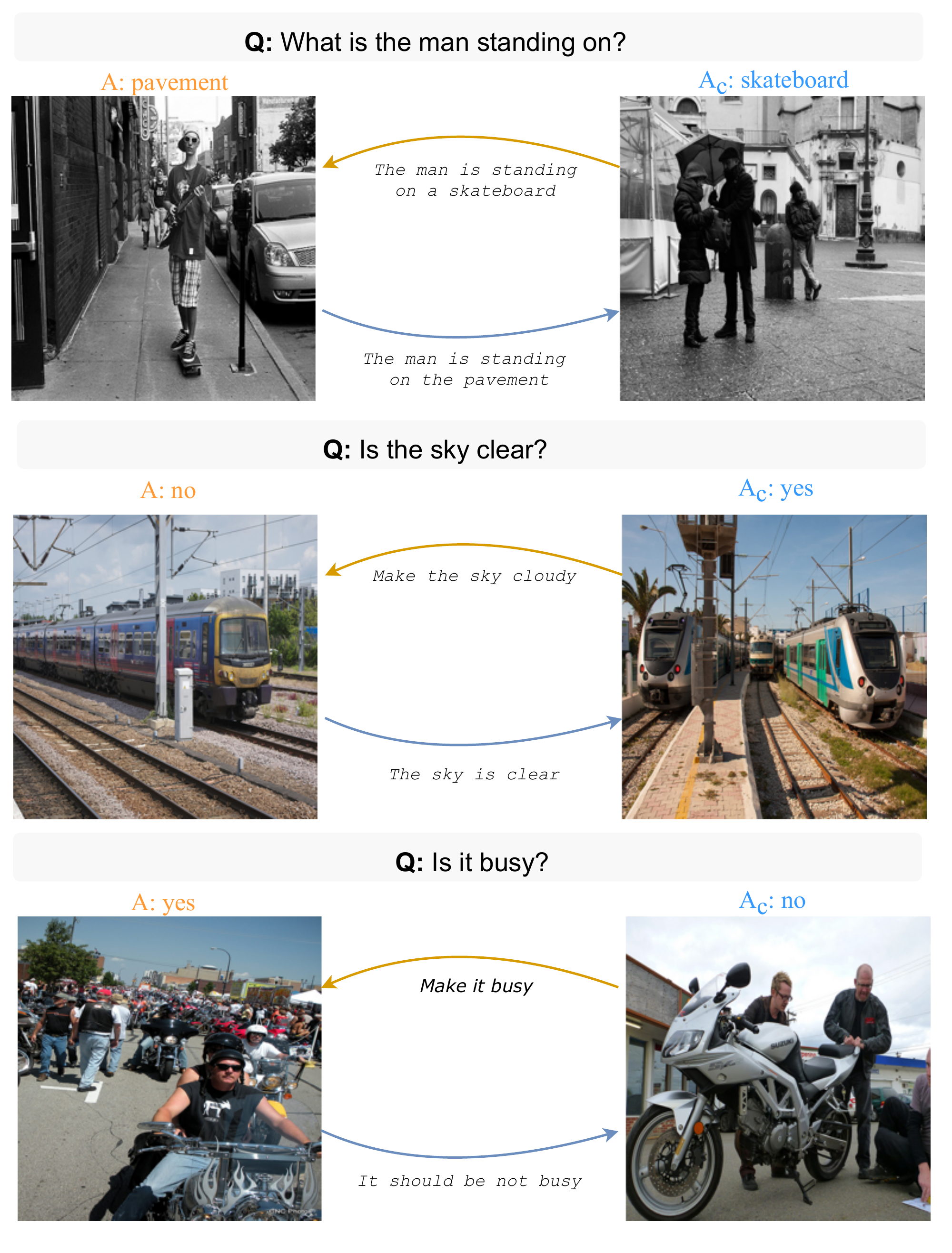}
   \caption{Creating \ourDS triplets from VQA2.0, using our Data Roaming approach. The question, associated answers together with the original and complimentary images are shown. Note the transition-texts automatically created by our approach.}
   \label{fig:vqa_to_lasco_1}
\end{figure*}

{\bf Few-shot Prompt:} Given the a triplet and its complementary, $(I,Q,A),(I_c,Q,A_c)\in \mathcal{D}$ where $\mathcal{D}:=$VQA2.0, we generate transition text from $I$ to $I_c$ based on $(Q, A_c)$. To use the few-shot ability of GPT-3, we provide it with a short {\color{Green}{description of the task}}, three manually {\color{Orange}{annotated examples}} and {\color{blue}a pair {$(Q, A_c)$} to rephrase}. An example for such input is presented below (used to generate the second example in \Cref{fig:vqa_to_lasco_1}):

{\it { \color{Green}{Rephrase the following texts:}}\textbackslash{}n
{\color{orange}{``Are there any humans visible in the photo? yes'' = ``Add some people to the photo''}}\textbackslash{}n
{\color{orange}{``Is there a square on the doors? yes'' = ``Find a door with a square on top of it''}}\textbackslash{}n
{\color{orange}{``What color is the man's tie? blue'' = ``Change the color of the tie of the man to be blue''}}\textbackslash{}n
{\color{blue}{``Is the sky cloudy? yes'' =}}}\\
This input results in the transition-text ``Make the sky cloudy''.

\begin{figure*}[t]
  \centering
  \includegraphics[width=1\textwidth,height=0.95\textheight]{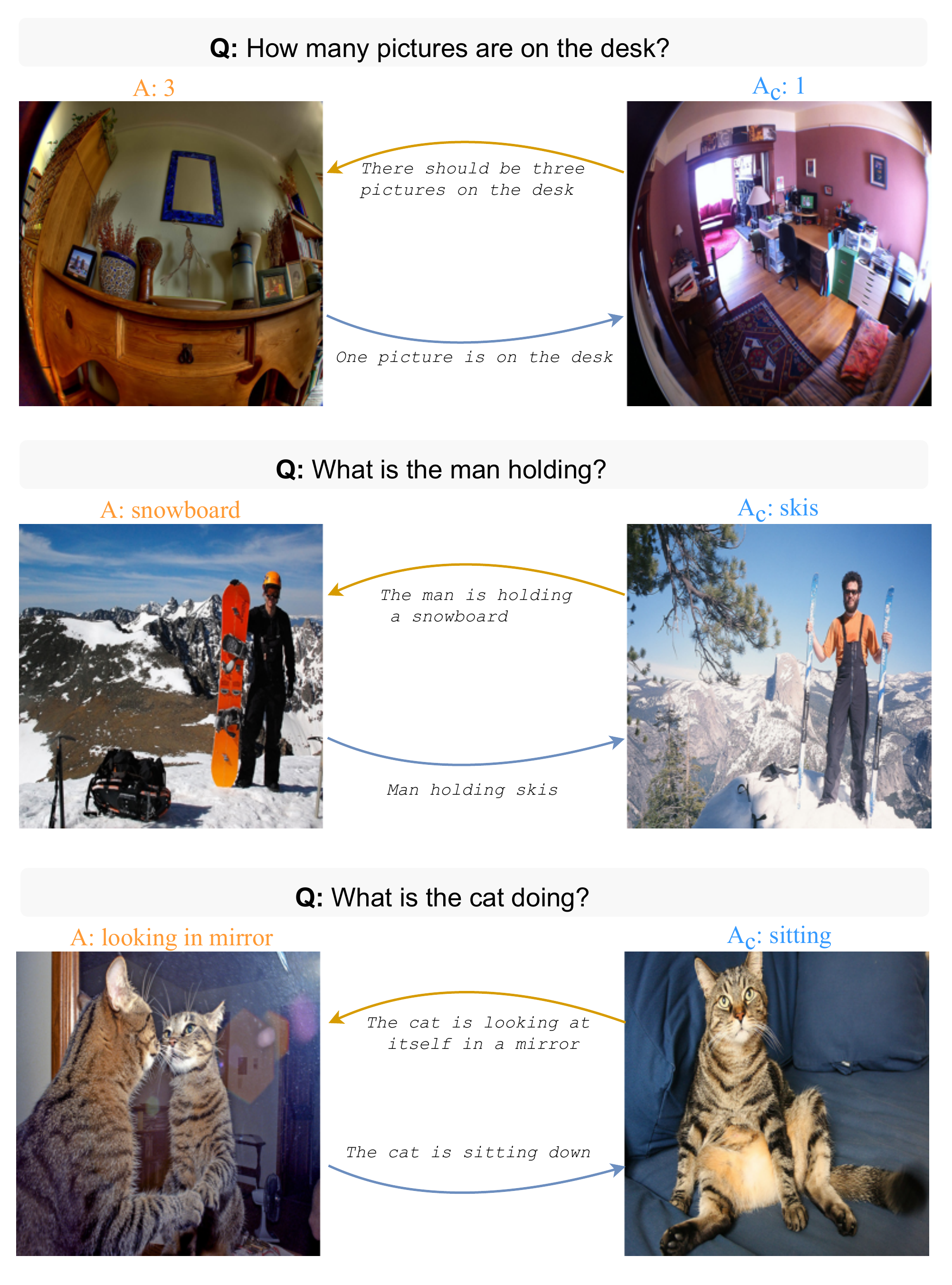}
   \caption{Creating \ourDS triplets from VQA2.0, using our Data Roaming approach. The question, associated answers together with the original and complimentary images are shown. Note the transition-texts automatically created by our approach.}
   \label{fig:vqa_to_lasco_2}
\end{figure*}

\begin{figure*}[t]
  \centering
  \includegraphics[width=1\textwidth,height=0.95\textheight]{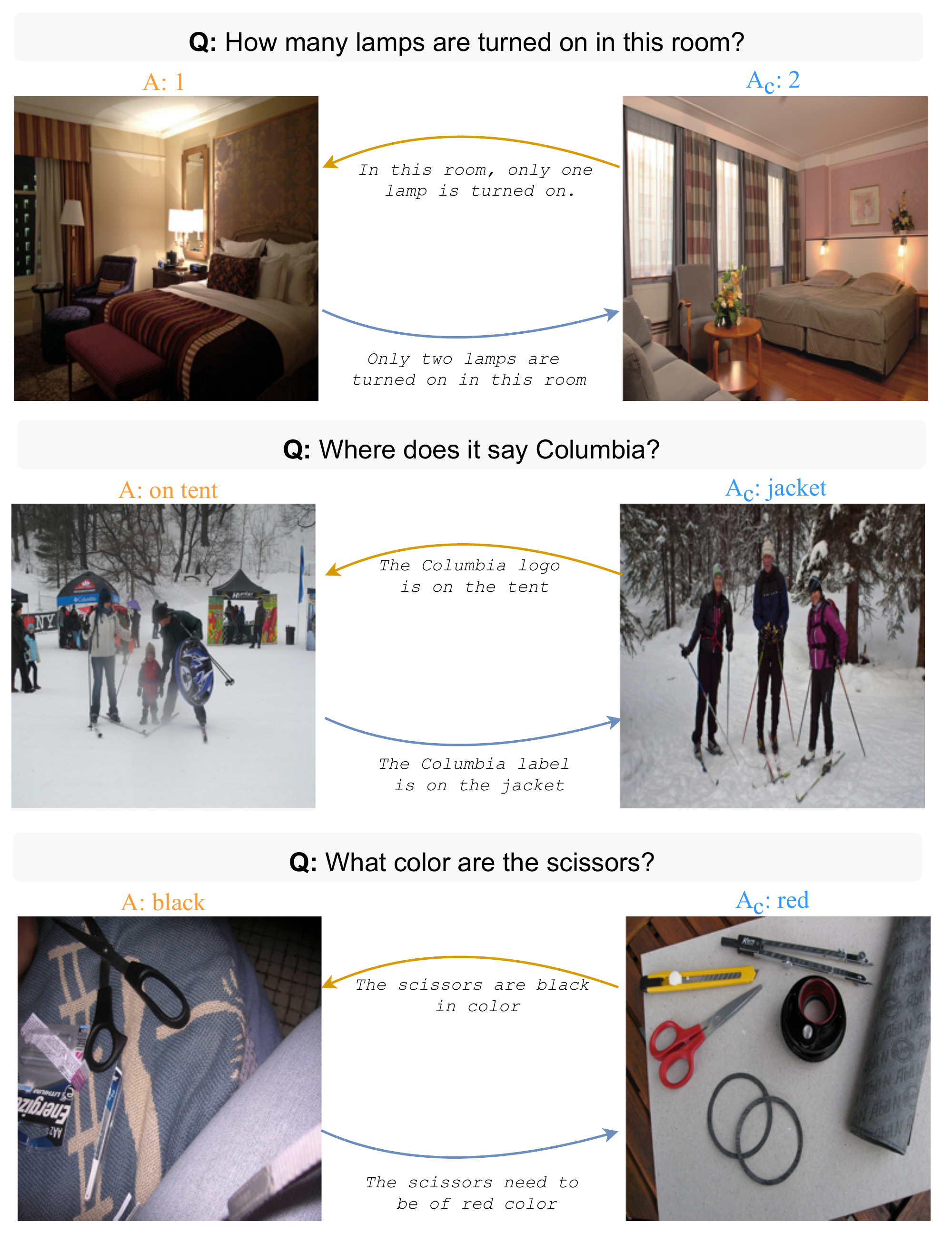}
   \caption{Creating \ourDS triplets from VQA2.0, using our Data Roaming approach. The question, associated answers together with the original and complimentary images are shown. Note the transition-texts automatically created by our approach.}
   \label{fig:vqa_to_lasco_3}
\end{figure*}

\begin{figure*}[t]
  \centering
  \includegraphics[width=1\textwidth,height=0.7\textheight]{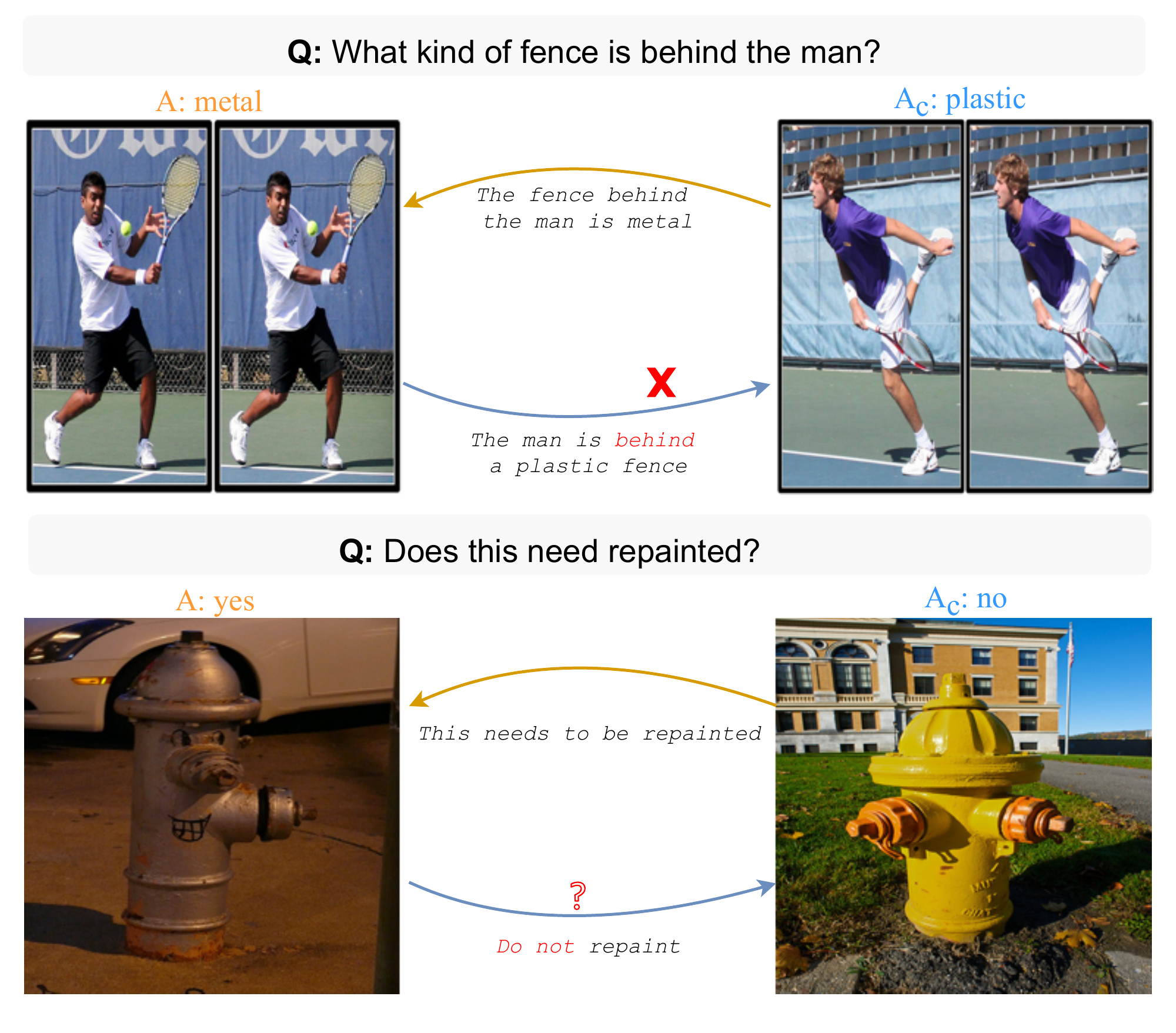}
   \caption{Arguably failure converting cases, while creating \ourDS triplets from VQA2.0, using our Data Roaming approach. The duplicated image at the top is from the source. First row shows a logic failure, since the fence is behind the man, and not the other way. Second row shows a questionable rephrasing.}
   \label{fig:vqa_to_lasco_fail}
\end{figure*}

Several examples with the associated images, determining \ourDS triplets are shown in  \Cref{fig:vqa_to_lasco_1,fig:vqa_to_lasco_2,fig:vqa_to_lasco_3}. Using the symmetry we show transition text in both directions (See Sec.~5 in the paper). We observe that generally the generated text correctly describes the change between the images. It is natural, coherent and grammatically correct (due to the capabilities of GPT-3). We further see in many examples that the rephrasing of GPT-3 is not limited to the vocabulary in the question and the answer. In some cases \eg, {\it Is the sky clear? No,} the model shows high level of understanding by using new words in the created transition text, as in here {\it ``Make the sky cloudy''} (second example in \cref{fig:vqa_to_lasco_1}).
However, there are also cases where GPT-3 fails to produce an appropriate transition-text. Two such examples are shown in \Cref{fig:vqa_to_lasco_fail}.

\subsection{CoIR User Study}
Here we provide more details regarding our user study that was separately conducted on \ourDS, FashionIQ and CIRR datasets. We sample $\sim300$ random CoIR triplets from the dataset, each one of them is composed of query-image, target-image and a transition-text. We present each triplet by showing the transition-text and the two images, query and target, side by side as left and right, respectively. We ask the users to rate each triplet just by positive/negative (``Good''/``Bad'') labels.
We crowd-sourced 30 students in the ages of 20-35, from The Hebrew University of Jerusalem, for the evaluation.
All individuals were asked to review the dataset annotations with the following instructions:

\begin{quotation}
``We want you to evaluate the following annotations for the 'Image retrieval' task:\\
{\bf The input}: image (left) + text\\
{\bf The Output}: image (right)\\

The target is to find the (right) image within a large database of images.\\
The query for search is the left image, serves as context, and the text is some modification/request.

How well can the target image on the {\bf right} be derived from the source image on the {\bf left}? If you find that the target image is adequately described by the source image and the required change in the text, please label it as `\textcolor{green}{Good}', otherwise as `\textcolor{red}{Bad}'. ''
\end{quotation}

\begin{figure}[t]
  \centering
  \includegraphics[width=1\linewidth]{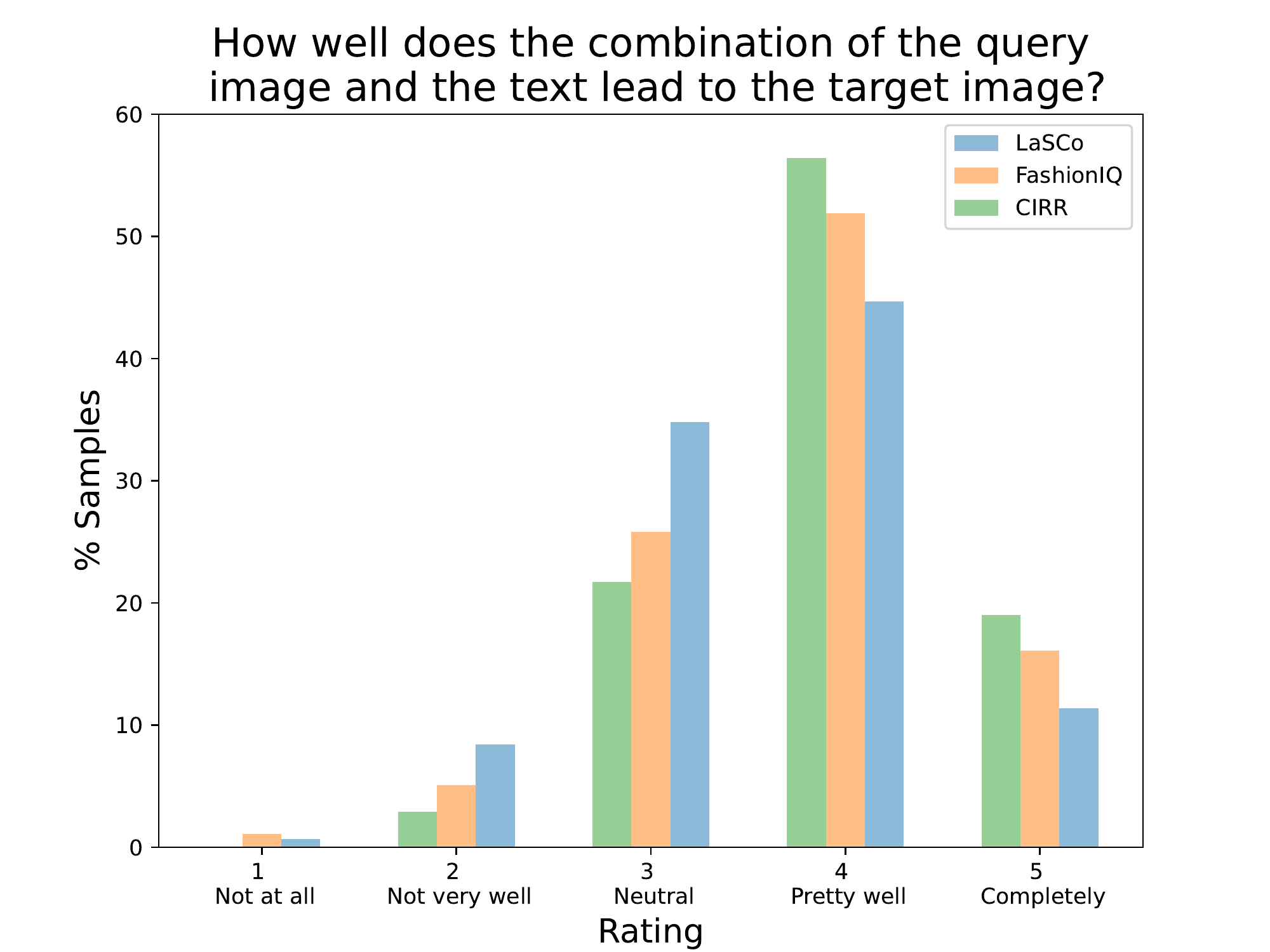}
   \caption{Users rating of CoIR annotations, conducted on 1000 random samples per dataset.}
   \label{fig:datasets_amt}
\end{figure}

{\bf User study validation:} In order to further validate the results of the user study described above, we performed a similar study on a larger scale using the Amazon Mechanical Turk (AMT) platform. Specifically, 1000 random samples from each dataset were rated by 3 different AMT workers, using a 1--5 rating scale (worst--best, respectively). A mean opinion score (MOS) was computed for each sample as the average of the three ratings. Rating comparison presented in \Cref{fig:datasets_amt}. 
%
Binarization of the ratings (considering 1,2 as `Bad', otherwise as `Good') yields a positive (Good) rate of 90.9\%, 93.8\% and 97.1\% for \ourDS, FashionIQ and CIRR, respectively. The corresponding MOS scores were 3.58, 3.77 and 3.92. The Overall (relative) gap between \ourDS and the other datasets is up to $7\%$. This further confirms that although \ourDS was generated using data roaming, the positive rate of the samples is nearly as high as for datasets that were manually annotated.

Raters were chosen from Amazon Mechanical Turk (AMT) with an HIT approval rate of at least 98\%, from the United States and United Kingdom. 

\section{Modality Redundancy}
\label{sec:modality_redundancy_supp}
\begin{figure}[t]
  \centering
  \includegraphics[width=1\columnwidth]{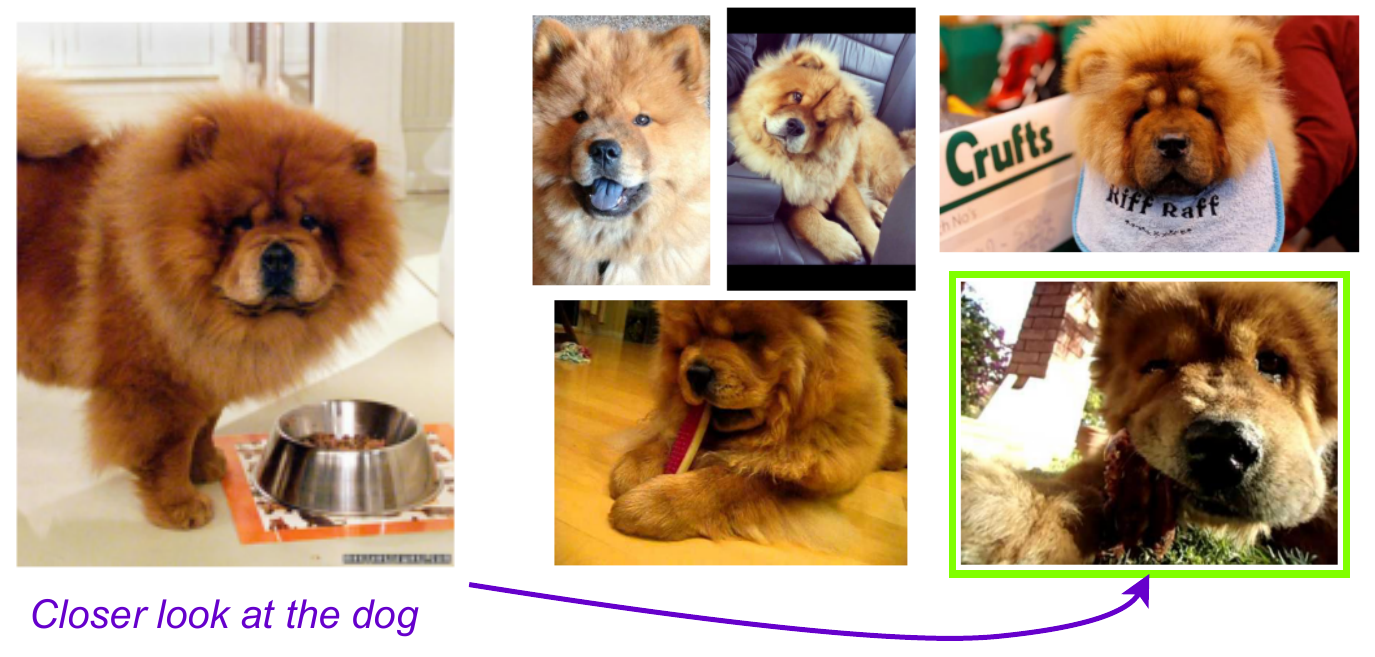}
   \caption{A CIRR query-target pair within its collected image sub-group. The transition-text is descriptive enough to reach the target-image (framed in green), within this sub-group.}
   \label{fig:cirr_problematic_ex}
\end{figure}
 Here we present more examples on the modality redundancy effect that we address in the paper (Sec.~5). 
 \Cref{fig:cirr_retrievals} shows examples from CIRR validation set that were retrieved by our \our.
 Note that in rows 1-6, the query texts are quite descriptive and likely adequate to retrieve the target, solely based on textual information. We put it to test by also applying \our(Text-Only) on these examples. Based on transition texts only, it successfully retrieved the target images at rank \#1, implying the redundancy of the ignored query images.
 A different example of modality redundancy is reflected in the ``Sub-group" experiments in CIRR. In these experiments a manual transition-text was created to map the query image (on the left) to the target (framed in green) while it was shuffled, by construction, in a sub-group of visually similar images. 
Despite the fact that query-text is not descriptive enough to retrieve the target image from the entire corpus (since it might refer to any kind of dog), it is sufficient to do so within the particular sub-group, due to the high visual similarity. The success \our(Text-Only) on this particular subtask reaching the highest $\text{Recall}_{\text{subset}}@K$ (cf. Table 3, in the main paper) can be related to this fact. 

\section{Ablation Study}
\label{sec:ablation_study}
In this section we show our ablation analysis also on other datasets, in addition to FashionIQ presented in the main paper.
\Cref{tab:cirr_ablations} shows further ablation results on the CIRR dataset. As we expected, our Reverse-Query objective is less crucial to the CIRR dataset, since the transition-texts are ``over-informative" for selecting the target images (as discussed in the main paper).
\begin{table}[ht]
\begin{center}
\resizebox{\columnwidth}{!}{%
\begin{tabular}{@{}lllllll@{}}
\toprule
 & R@1 & R@5 & R@10 & R@50 & R@100 & R@500 \\ \midrule
LF-BLIP & 21.96 & 50.16 & 64.05 & 85.67 & 90.62 & 98.04 \\
No-RQ & 47.50 & 80.00 & 88.85 & 97.35 & 98.71 & 99.86 \\
Contrastive Loss & 48.00 & 80.08 & {\bf 89.07} & {\bf 97.56} & {\bf 98.92} & 99.86 \\
Freeze ViT & {\bf 48.07} & 79.60 & 88.52 & 97.27 & 98.59 & 99.78\\
\our & { 47.96} & {\bf 80.65} & 88.88 & 97.46 & 98.78 & 99.81 \\
\bottomrule
\end{tabular}
}
\end{center}
\caption{Ablation study on CIRR validation dataset.}
\label{tab:cirr_ablations}
\end{table}

\subsection{Batch Size Ablation}
Here, we show ablation results on the batch size in CASE training. \Cref{tab:fiq_ablations_bs} presents the results with different batch sizes conducted on \fashioniq dataset. The optimal batch size is around the range 512-1024.
\begin{table}[ht]
\begin{center}
\resizebox{\columnwidth}{!}{%
\begin{tabular}{@{}l|llllll@{}}
\toprule
Size & R@1 & R@5 & R@10 & R@50 & R@100 & R@500 \\ \midrule
32 &  0.91 & 3.69 & 5.85 & 15.87 & 22.86 & 47.21 \\
64 &  18.1 & 36.24 & 45.51 & 68.58 & 77.14 & 92.5 \\
128 &  18.62 & 37.3 & 46.33 & 69.13 & 77.74 & 92.29 \\
256 &  19.4 & 38.25 & 47.64 & {70.28} & {78.37} & \textbf{92.64} \\
512 &  19.4 & {38.58} & \textbf{48.37} & 70.2 & \textbf{78.44} & {92.55} \\
1024 &  \textbf{20.46} & \textbf{39.03} & {48.24} & \textbf{70.31} & 78.26 & 92.34 \\
2048 &  {20.1} & 38.26 & 47.71 & 69.78 & 78.01 & 92.24 \\
4096 &  20.1 & 37.85 & 47.16 & 68.7 & 76.98 & 91.26 \\
\end{tabular}

}
\end{center}
\caption{Batch size ablation study on FashionIQ dataset. }
\label{tab:fiq_ablations_bs}
\end{table}

\section{Explainability}
\label{sec:explainability}
We also present an explainability tool for our model for analysis and better understanding its reasoning pattern. To this end, we use two existing methods to generate a visual and textual ``reasoning", during the retrieval.  

{\bf Visual modality:} Following \cite{vasu2021explainableCBIR}, we slide a black window (mask) over the query-image and feed it to our model. For each masked query-image, we calculate the cosine distance of the model's output to the ground truth target, in the embedding space. We overlay the result as a heat-map over the query image. ``Hotter'' (tending to red) regions indicate increased importance in retrieval (since masking them results in a larger distance).

\begin{figure}[t]
\begin{subfigure}[b]{1\columnwidth}
         \centering
         \includegraphics[width=\textwidth]{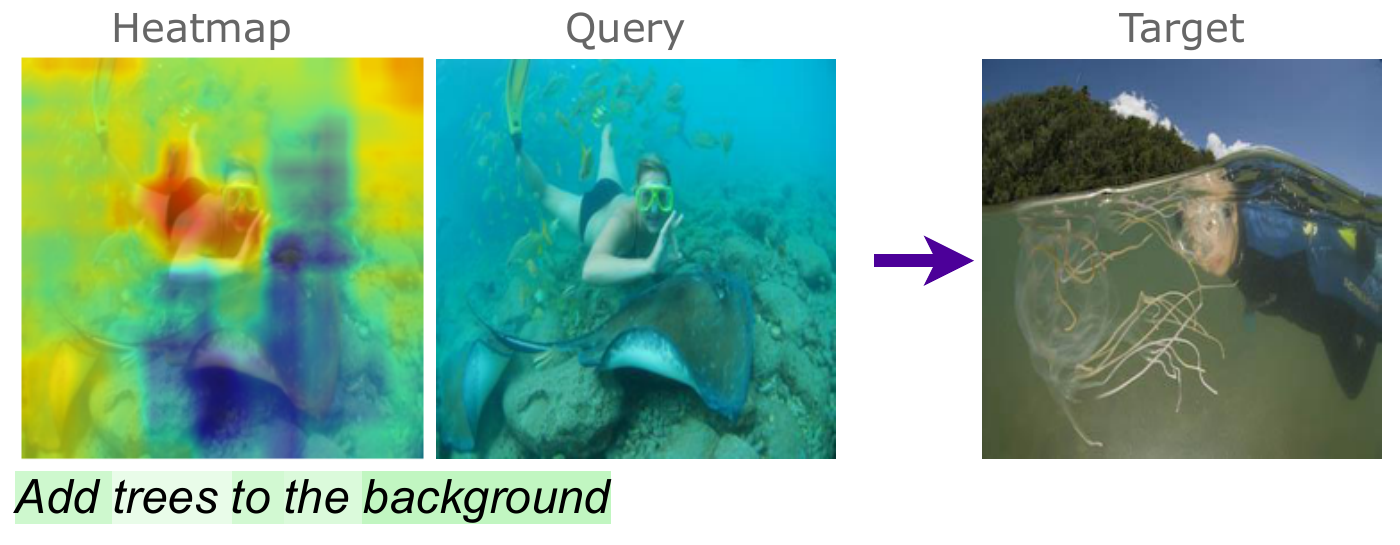}
         \caption{Preserving human in target image}
         \label{fig:cirr_exp_1}
     \end{subfigure}
     \vfill
\begin{subfigure}[b]{1\columnwidth}
         \centering
         \includegraphics[width=\textwidth]{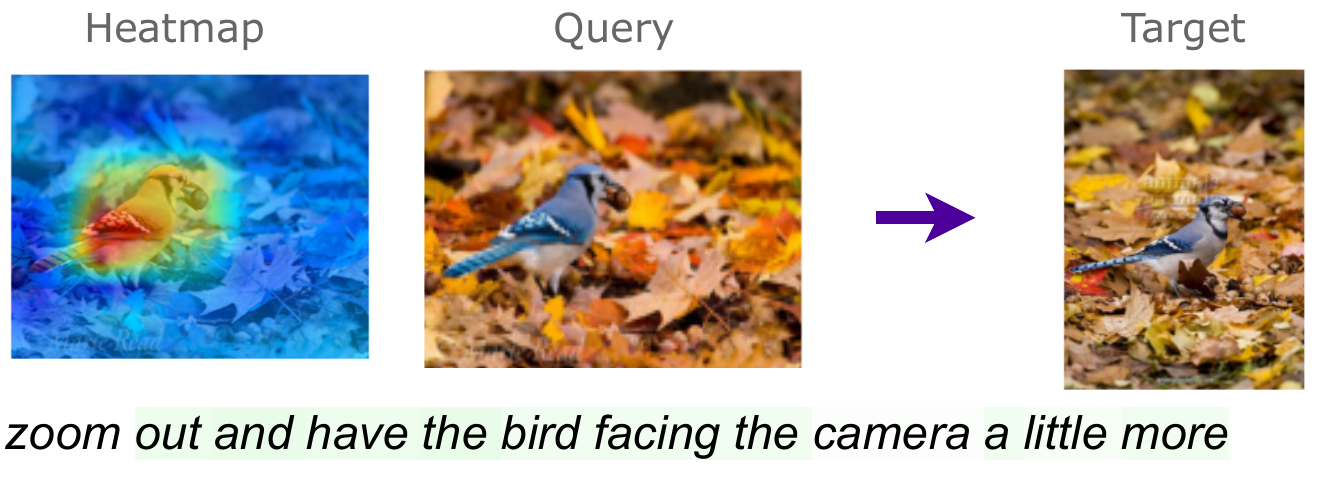}
         \caption{ }
         \label{fig:cirr_exp_2}
     \end{subfigure}
     \vfill
\begin{subfigure}[b]{1\columnwidth}
     \centering
     \includegraphics[width=\textwidth]{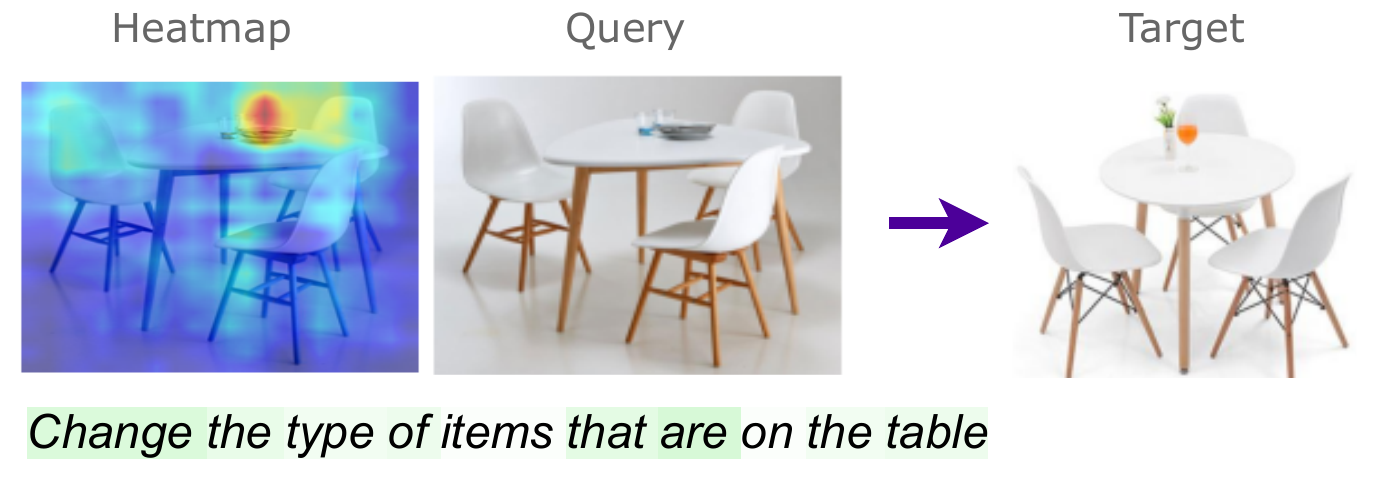}
     \caption{ }
     \label{fig:cirr_exp_4}
 \end{subfigure}
 \vfill
\begin{subfigure}[b]{1\columnwidth}
     \centering
     \includegraphics[width=\textwidth]{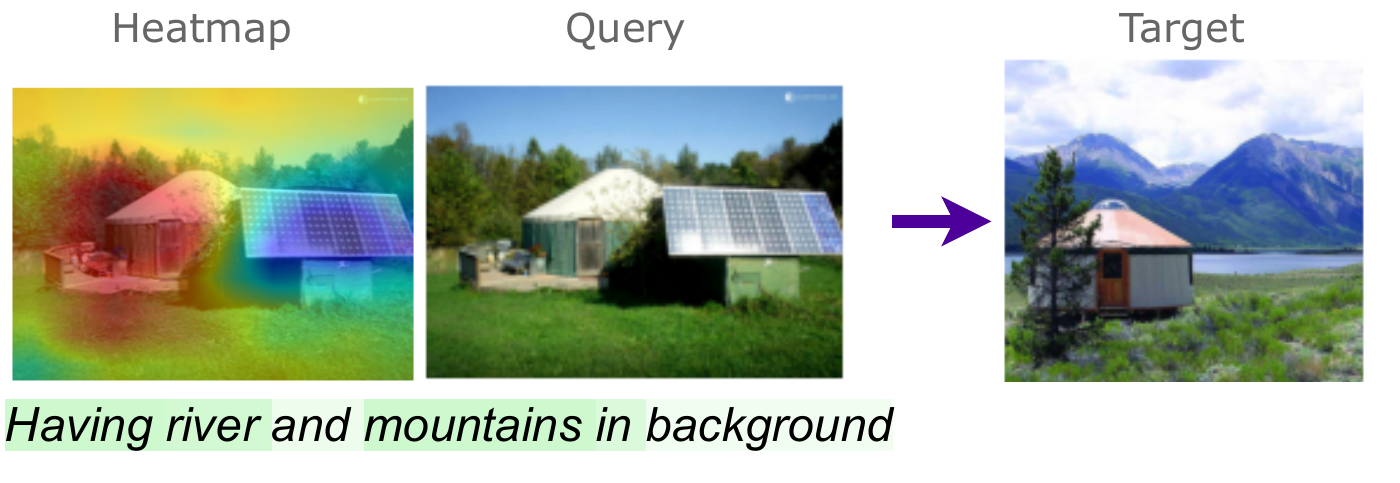}
     \caption{Preserving same looking house}
     \label{fig:cirr_exp_5}
 \end{subfigure}
 \vfill
\begin{subfigure}[b]{1\columnwidth}
     \centering
     \includegraphics[width=\textwidth]{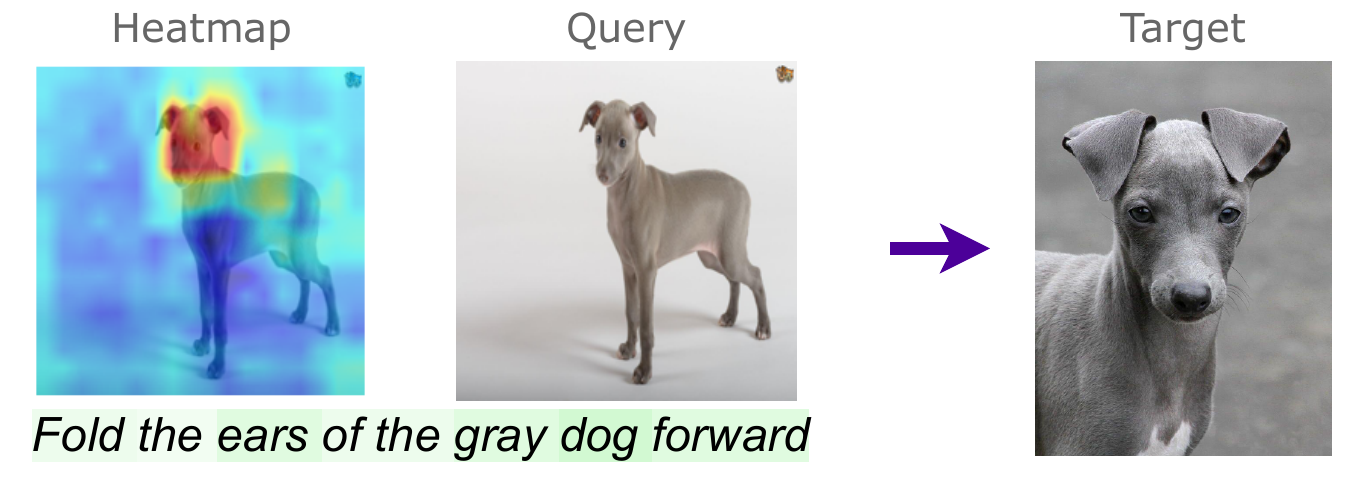}
     \caption{ }
     \label{fig:cirr_exp_6}
 \end{subfigure}
 \vfill
   \caption{Explainability heatmaps for \coir task, tested on our model (CASE). Model's attention is indicated by 'jet' heatmap over the image and background highlight over the transition-text. CASE combines information from perceptually reasonable parts of the image and text to reach the target.}
   \label{fig:cirr_exp}
 
\end{figure}

{\bf Text modality:}
For visualizing the relevance of text tokens, we calculate the gradients over attention maps in the shift-encoder, w.r.t the desired target, using the multi-modal visualization method as suggested in \cite{Chefer_2021_ICCV}). 

These two maps create a visualization on the attended regions and textual tokens in the query. Note that, we expect the model to focus on {\it complementary} information between the query-image and transition text. It needs to ``see" the gist or the object in the image and ``understatnd" the change or a specification from the transition-text. 

We demonstrate this explainability method on a few examples in \Cref{fig:cirr_exp}. In \Cref{fig:cirr_exp_1} a diver is shown. The transition text asks for change in the background. As observed, the model attends the diver and the sea in the image (not specified in text), as well as the words such as `Add' and `background'. \Cref{fig:cirr_exp_6} shows another example with a dog, with the transition-text asking to fold its ears forward. Our model attends the head of the dog in the image, probably identifying the dog's breed and ears and corresponding to important cues in the text, \eg `ears' and `forward', to retrieve the correct target image.

\end{document}